%% file: branchGAN.tex
\documentclass[oribibl, twocolumn]{llncs}  
\usepackage{graphicx}
\usepackage [pagewise]{lineno}

\usepackage{amsmath,amssymb} 
\usepackage{color}
\usepackage[width=180mm,left=18mm,paperwidth=215mm,height=240mm,top=18mm,paperheight=279mm]{geometry}
\usepackage{hyperref} 

\usepackage[T1]{fontenc}
\usepackage[utf8]{inputenc}
\usepackage{authblk}

\usepackage{color}

\usepackage{titles}
\usepackage{etoolbox} 
\usepackage[dvipsnames]{xcolor}

\DeclareMathOperator{\HOG}{HOG}
\DeclareMathOperator{\VBS}{VBS}
\DeclareMathOperator{\DFT}{DFT}

\begin{document}

%
\pagestyle{empty}
\title{BSD-GAN: Branched Generative Adversarial \\ Network for Scale-Disentangled \\ Representation Learning and Image Synthesis}

\author{ Zili Yi\textsuperscript{1,2} 
\and  Zhiqin Chen\textsuperscript{1} 
\and Hao Cai\textsuperscript{2} 
\and Wendong Mao\textsuperscript{2} 
\and Minglun Gong \textsuperscript{3}
\and Hao Zhang \textsuperscript{1}} 

\institute{
 \textsuperscript{1} Simon Fraser University, Canada \\ 
 \textsuperscript{2} Memorial University of Newfoundland, Canada \\
 \textsuperscript{3} University of Guelph, Canada}




\twocolumn[
  \begin{@twocolumnfalse}
	\maketitle
\begin{abstract}

We introduce {\em BSD-GAN\/}, a novel {\em multi-branch\/} and {\em scale-disentangled} training method which enables unconditional Generative Adversarial Networks (GANs) to learn image representations at {\em multiple scales\/}, benefiting a wide range of generation and editing tasks. The key feature of BSD-GAN is that it is trained in multiple branches, progressively covering both the breadth and depth of the network, as resolutions of the training images increase to reveal finer-scale features. Specifically, each noise vector, as input to the generator network of BSD-GAN, is {\em deliberately \/} split into several sub-vectors, each corresponding to, and is trained to learn, image representations at a particular scale. During training, we progressively ``de-freeze'' the sub-vectors, one at a time, as a new set of higher-resolution images is employed for training and more network layers are added. A consequence of such an explicit sub-vector designation is that we can {\em directly\/} manipulate and even combine latent (sub-vector) codes which model different feature scales.
Extensive experiments demonstrate the effectiveness of our training method in {\em scale-disentangled\/} learning of image representations and synthesis of novel image contents, without any extra labels and without compromising quality of the synthesized high-resolution images. We further demonstrate several image generation and manipulation applications enabled or improved by BSD-GAN. Source codes are available at \href{https://github.com/duxingren14/BSD-GAN}{https://github.com/duxingren14/BSD-GAN}.
\end{abstract}
  \end{@twocolumnfalse}
]

\input{introduction}

\input{related}
\input{method}

\input{experiment}

\input{conclusion}

\bibliographystyle{splncs}
\bibliography{reference}
\end{document}

%% file: introduction.tex
\section{Introduction}
\label{sec:intro}

Unconditional Generative Adversarial Networks (GANs)~\cite{goodfellow2014generative} have been intensively studied for unsupervised learning and data synthesis. Compared to their conditional counterparts~\cite{mirza2014conditional,yan2016attribute2image,odena2016conditional,reed2016generative,zhang2017stackgan,isola2017image,zhu2017unpaired,yi2017dualgan}, 
unconditional GANs place less burden on the training data but are less steerable at the same time. In an unconditional GAN, a well-trained generator could synthesize novel data by sampling a random noise vector from the learned manifold as input and altering values 
``parameterizing'' the dimensions of the manifold. 
However, this synthesis process is typically uncontrollable and counterintuitive, since we have little understanding of how each manifold 
dimension impacts the synthesized output. 

\begin{figure}[!t]
\centering
\includegraphics[width=.95\linewidth]{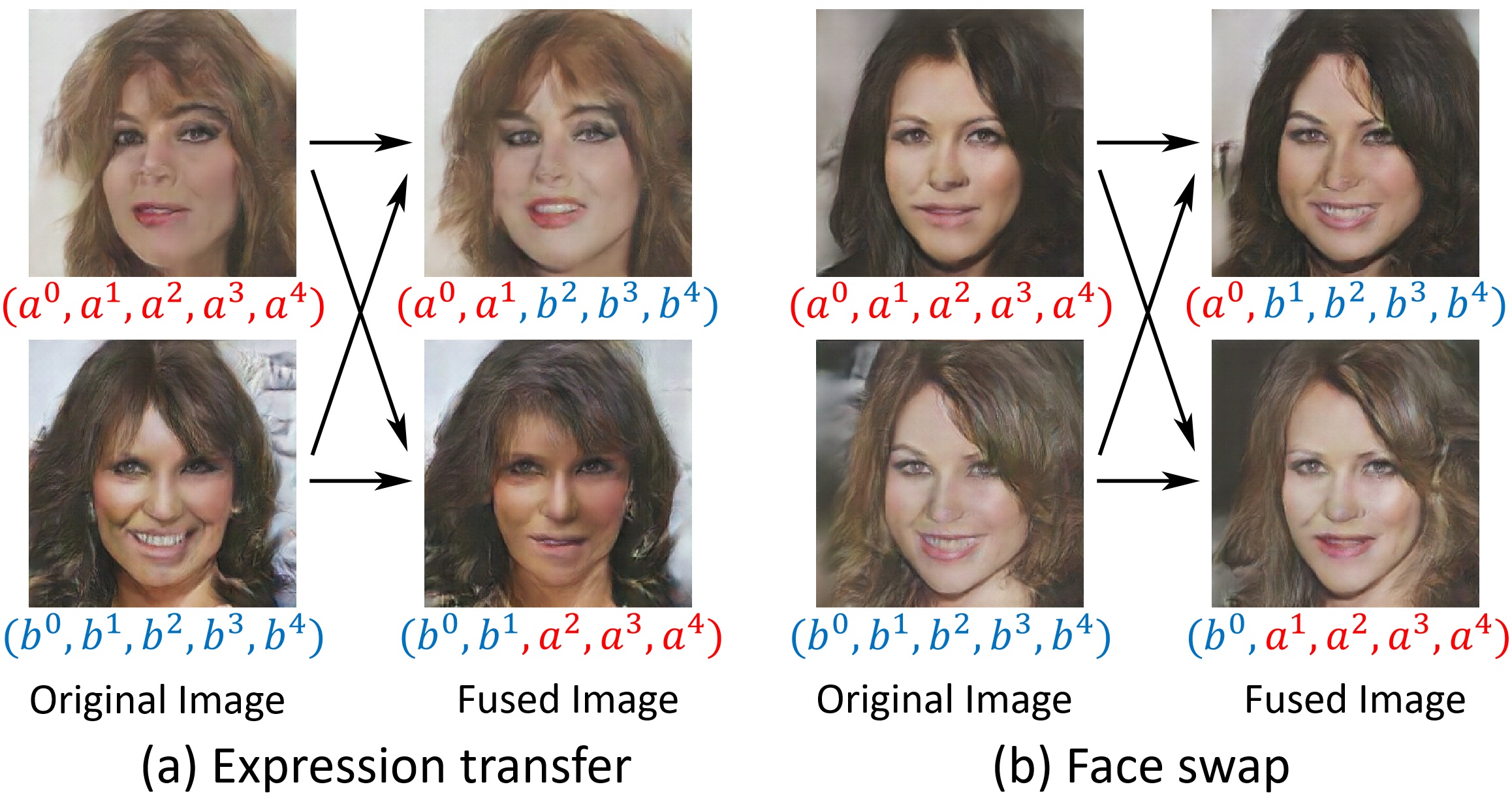}
\caption{{\em Cross-scale image fusion\/} enabled by our multi-branch and scale-disentangled training of unconditioned GANs, where each noise vector for image generation is split/branched into five sub-vectors that are trained to learn image representations at different scales. The fusion is achieved by directly combining coarse-scale features in one image with finer-scale features from another. Please note that $\mathbf{x}^0 (\mathbf{x} \in \{\mathbf{a},\mathbf{b}\})$ encodes image-wide structures and $\mathbf{x}^t (t \in \{1,2,3,4\})$ encodes increasingly finer-scale features. Given a pair of images, we synthesize new images by cross-combining coarse-scale structures and fine-scale features of the two input images, accomplishing expression transfer (a) and face swap (b), respectively.} 
\vspace{-15pt}
\label{fig:teaser}
\end{figure}

For manifold learning of images or other visual forms, the notion of {\em feature scales\/} is of great importance. The ability
to learn multi-scale or scale-invariant features often leads to a deeper and richer understanding of representations and distributions
of images. In the last few years, {\em scale-aware unconditional GANs\/} have been developed, e.g., StackGAN~\cite{zhang2017stackgan}, LPGAN~\cite{denton2015deep} and PGGAN~\cite{karras2017progressive}, where correlated GANs are trained in a coarse-to-fine manner, using lower- and then higher-resolution images, with the goal of improving the quality of the final full-resolution images. However, factors which impact image features at various scales remain entangled in PGGAN networks. In StackGAN and LPGAN, factors (dropout layer or noisy perturbation) were added at various scales to learn scale-independent image features, though they are neither explicit nor controllable.
Under the setting of unconditional or conditional GANs, several recent works, including InfoGAN~\cite{chen2016infogan}, DNA-GAN~\cite{xiao2018dnagan}, SD-GAN~\cite{donahue2018}, aim to disentangle the latent codes, which correspond to different image {\em attributes\/}, rather than image feature scales. 

\begin{figure*}[!t]
\centering
\includegraphics[width=0.8\linewidth]{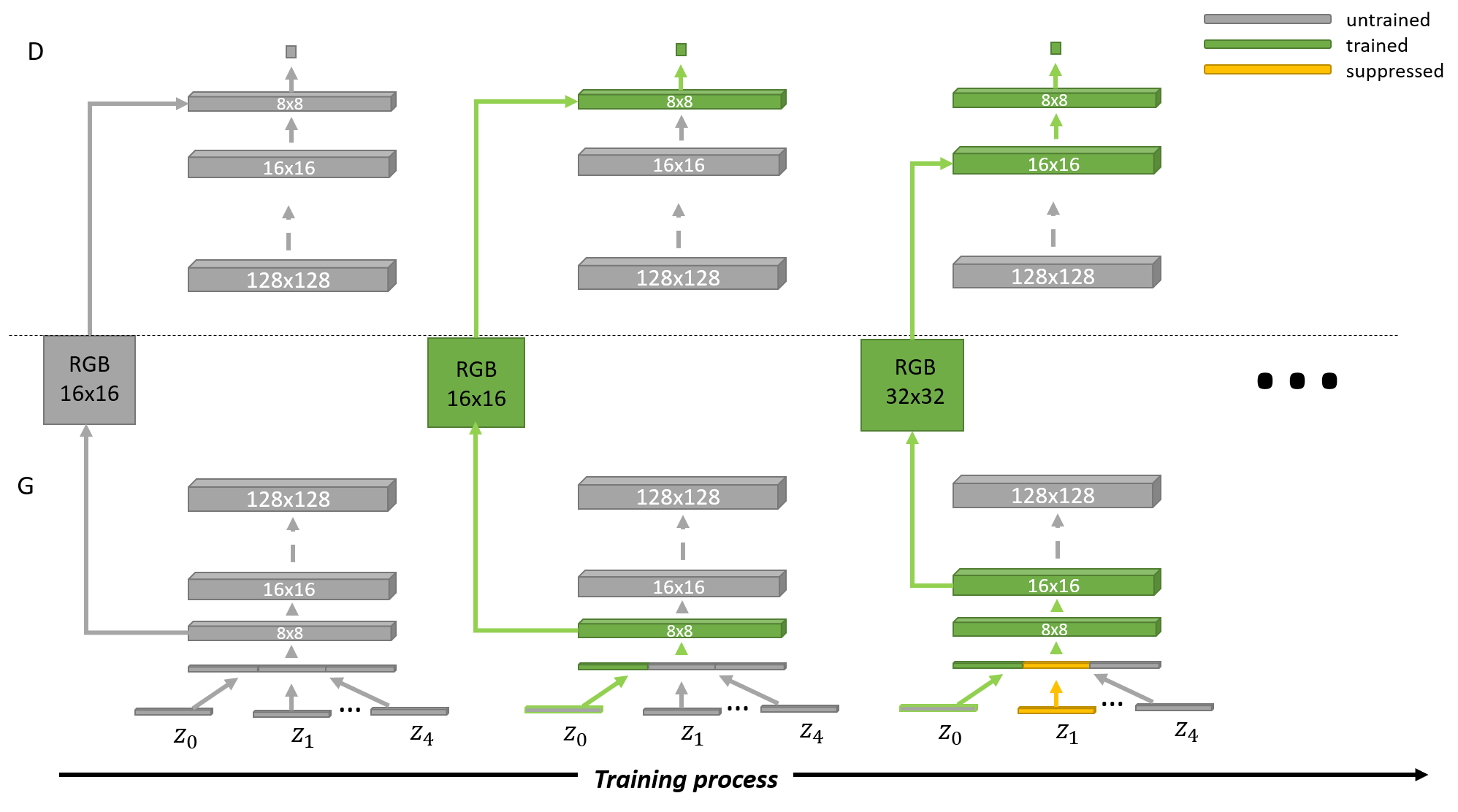}
\vspace{-10pt}
\caption{{\em Training pipeline of BSD-GAN.\/} We start with both the generator (G) and discriminator (D) having a low spatial resolution. During the first training period, we feed $\mathbf{z}^0$ with random vectors of uniform distribution and $\mathbf{z}^t$ ( $t > 0$) with zero vectors $\mathbf{0}$, and make those linear layers corresponding to $\mathbf{z}^t$ ( $t > 0$) untrainable. As the training advances, we incrementally add layers to G and D, thus increasing the spatial resolution of the generated images. Meanwhile, we ``de-freeze'' more $\mathbf{z}$ vectors for training by feeding them with non-zero uniform-random vectors. This process is repeated until the target resolution is reached. During training, ``branch suppression'' (see Section~\ref{sec:method} for details) happens as $\mathbf{z}^0$ has well encoded large-scale structures and will maintain its dominance in coarse-level encoding. When $\mathbf{z}^1$ is de-frozen, it is suppressed in terms of coarse-level encoding but has the chance to encode finer-scale features.}
\vspace{-10pt}
\label{fig:framework}
\end{figure*}

In this paper, we introduce a novel training method which enables unconditioned GAN generators to learn image representations in a {\em scale-disentangled\/} manner, aiming to improve the controllability of image synthesis and editing. The key novel feature of our learning paradigm is that each noise vector, as input to the generator, is {\em explicitly\/} split or branched into a prescribed number of sub-vectors, e.g., 5 for learning $256^2$ images and 6 for $512^2$ images, where each sub-vector corresponds to, and is trained to learn, image representations at a particular scale, without requiring extra labels.

At the high level, our learning method employs the standard GAN framework, which comes with an unconditioned generator and a discriminator, and follows the standard GAN training paradigm as described in~\cite{goodfellow2014generative,radford2015unsupervised}. To achieve scale-disentangled learning, our network is trained {\em progressively\/}, bearing some similarity to Karras et al.~\cite{karras2017progressive}. However, instead of progressing only on network depth (adding network layers as the resolutions of training images increase), our network training also progresses over the network {\em width\/} by progressively activating sub-vectors that correspond to different feature scales; see Figure~\ref{fig:framework}. As a result, we explicitly designate dimensions of the image manifold to different image scales, leading to scale-specific ``{\em training branches\/}''. We call our branched and scale-disentangled network BSD-GAN.

A consequence of the sub-vector designation in BSD-GAN is that we can {\em directly\/} manipulate and even combine latent (sub-vector) codes modeling different feature scales, leading to novel applications of unsupervised adversarial learning that were not possible before. 
Figure~\ref{fig:teaser} shows an example of a novel application, {\em cross-scale image fusion\/}, where we intentionally synthesize an 
image by integrating the coarse-scale features of one image with finer-scale features of another.
We tested our novel training method on several high-quality image datasets to verify its effectiveness in learning scale-disentangled image representations, compared to alternative GAN training schemes. We also demonstrate improved performance of iGAN~\cite{zhu2016generative}, for interactive image generation, when the generator is trained by BSD-GAN in place of DCGAN. 


In summary, the main contributions of our paper include:
\begin{itemize}
\item A novel training method, BSD-GAN, that enables unconditioned generators to learn image representations in a {\em scale-disentangled\/} manner, without any extra labels.
\item The design of novel applications (cross-scale image fusion and coarse-to-fine image synthesis) enabled by the scale-disentangled latent codes of BSD-GAN, as well as improvements over existing applications (iGAN).
\item A high-resolution ($800 \times 600$) image dataset to stimulate research on image synthesis and manifold learning.
\end{itemize}

What had originally motivated BSD-GAN and what made it work effectively is a phenomenon that we observed during our experiments with multi-branch data generators. Specifically, we found that when multiple noise vectors, with their respective training branches are at play, GAN training typically results in one dominant branch while the other branches are either fully or partially {\em suppressed\/}. In other words, the already-trained weights (branches) will have priority in maintaining their role in encoding the image structures that are already encoded and suppress the other branches. We refer to this phenomenon as ``{\em branch suppression\/}'' and believe that this observation may offer insight in other contexts of training multi-branch convolutional and generative networks.

%% file: related.tex
\section{Related work}
\label{sec:related}

\subsection{Disentangled image representation and generation}

One of the most important objectives of representation learning is to {\em disentangle\/} the underlying
factors of variation in the data~\cite{bengio2013}. In the realm of disentangled image representation 
learning and generation, most works have focused on disentangling image attributes~\cite{chen2016infogan,xiao2018dnagan,donahue2018}
or style-content separation~\cite{MUNIT,DI2I,zhou2019branch}. In the latter case, the networks are typically
designed to learn a share latent space representing content but separate attribute spaces to encode
different styles, where manipulation in the style spaces allows controllability during image generation.
Compared to these works, the novelty of BSD-GAN is reflected by its progressive training over both the
depth and width of the network for scale disentanglement, with a scale-aware 
sub-vector designation.

\subsection{Multi-scale image representation}

An inherent property of visual objects is that they only exist as meaningful entities over certain ranges of scales in an image. How to describe image structures at multiple scales remains an essential and challenging problem in image analysis, processing, compression, and synthesis. Early methods for multi-scale image representing such as Discrete Fourier/Cosine Transforms (DFT/DCT)~\cite{ahmed1974discrete} and Discrete Wavelet Transformation (DWT)~\cite{shapiro1993embedded} are widely used in decomposing small-scale details and large-scale structures. In our paper, DFT is employed as a metric for scale disentanglement.


Another scale-independent representation of images is the layer activations of a well-trained Convolutional Neural Network (CNN)~\cite{lecun1998gradient,krizhevsky2012imagenet,simonyan2014very}. In a CNN, top activation layers roughly represent large-scale image structures such as objects and scenes, while bottom activations represent small-scale details such as edges, colors, or textures. Other than CNNs, stacked models such as Deep Belief Network (DBN)~\cite{hinton2006fast}, Stacked AutoEncoders (SAE)~\cite{vincent2010stacked,vincent2008extracting} and multi-scale sparse Coding~\cite{sallee2003learning} can also retrieve multi-scale representations of images, though their effectiveness could be limited.

Concurrent to our work, in StyleGAN, Karras et al.~\cite{karras2019style} developed a novel style-based generator for unsupervised separation of high-level attributes and stochastic variation in the generated images, enabling intuitive, scale-specific control of the image synthesis task. Inspired by prior works from the style transfer literature, they achieve scale-disentangled representation learning of images by refactoring the architecture of the generator, rather than transforming the way of training, as in BSD-GAN. While both works are built on the idea of progressively growing GANs~\cite{karras2017progressive}, our network architecture and training pipeline are considerably simpler --- it is a natural extension of progressive training over the network depth by introducing the novel idea of scale-specific progressive training over the designated sub-vectors.


\subsection{Scale-aware image synthesis}

Coarse-to-fine image synthesis has been explored in StackGAN~\cite{zhang2017stackgan}, LPGAN~\cite{denton2015deep}, and PGGAN~\cite{karras2017progressive}, 
as discussed in Section~\ref{sec:intro}. The goal of these methods is to synthesize higher-quality images, rather than to learn multi-scale image manifolds. 
We extend the idea of progressive training from progressively adding layers to progressively growing both layers and branches. 
In addition, instead of training multiple GANs~\cite{zhang2017stackgan,denton2015deep}, our method trains only one GAN.

\subsection{Controllability in image synthesis}

One of the most frequently adopted approaches to improving the controllability of image generation is conditional modeling. Conditional or semi-conditional GANs can condition the image synthesis task on image attributes~\cite{yan2016attribute2image,donahue2018,xiao2018dnagan, chen2016infogan}, classes~\cite{odena2016conditional}, input texts~\cite{reed2016generative,zhang2017stackgan}, or images~\cite{isola2017image,zhu2017unpaired,yi2017dualgan}. These methods either require extra labels or paired training images, or need strict inherent relations between the priors and the outputs. Our method, as a type of unconditional GAN, conditions image generation on random noise of uniform distribution and does not require any extra labels or priors.

For unconditional GANs, the method known as iGAN~\cite{zhu2016generative} provides a way for users to synthesize or manipulate realistic images in a controllable way. In iGAN, users can add certain constraints on the appearance of desired images (e.g., by drawing sketches, adding color strokes, or setting an exemplar image), while the latent code is then optimized to satisfy these user constraints. Nonetheless, a sophisticated optimization method is required, as gradient descent is particularly vulnerable to local minima. We demonstrate through experiments that the scale-disentangled latent spaces learned by BSD-GAN can help the original iGAN (with DCGAN) avoid falling into local minima and hence boost its performance.

%% file: method.tex
\section{Method}
\label{sec:method}

\subsection{Branched and scale-disentangled training}
\label{sect:framework}

\begin{figure}[!t]
\centering\includegraphics[width=\linewidth]{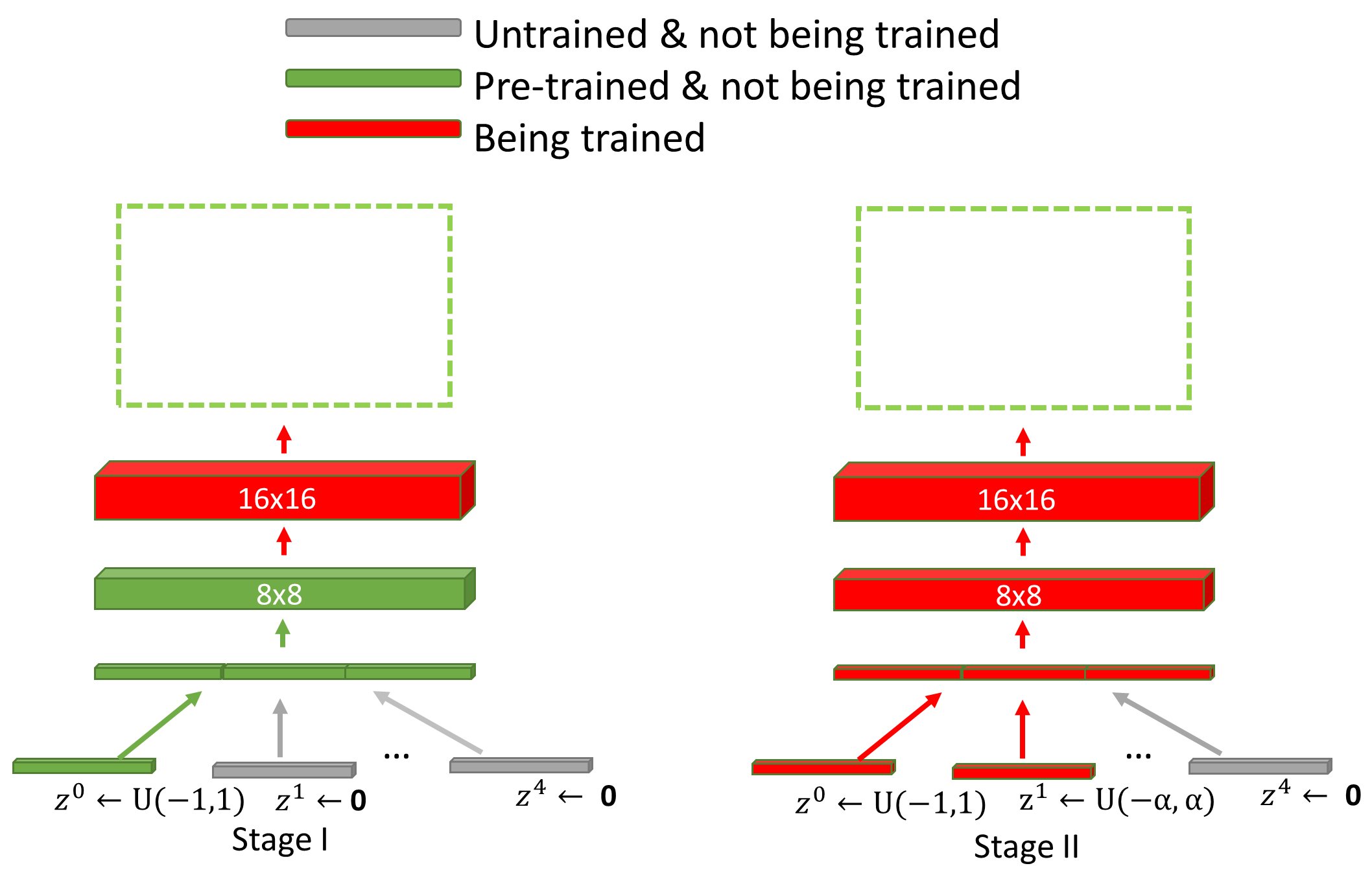}
\caption{{\em Training sub-procedure at one scale level.\/} After a new layer is added to the generator, we first only train the last layer of the generator while holding other layers untrainable (Stage I). After the last layer is well-trained, we then de-freeze all pre-trained layers (the branches in green) plus the newly-added branch for training (Stage II). To avoid ``sudden shock'' to the well-trained layers, we feed the newly de-frozen $\mathbf{z}$ sub-vector with a random vector following uniform distribution $U(-\alpha,\alpha)$, where $\alpha$ increases smoothly from $0.0$ to $1.0$ throughout Stage II.} \label{fig:framework2}
\end{figure}

Figure~\ref{fig:framework} visually describes the training pipeline of BSD-GAN. Specifically, our branched and scale-disentangled training starts only on the sub-vector corresponding to the coarsest level features, using the lowest-resolution images, while keeping the other sub-vectors ``frozen''. Then we progressively ``de-freeze'' the sub-vectors, one at a time, as a set of higher-resolution images is employed for training and more network layers are added. When training for a finer scale, the network weights learned from the previous coarser scales are used for initialization. Note that after training, these weights are often changed to adapt to the new training data.


Branch suppression does happen to the proposed branched training method. When de-freezing one sub-vector during progressive training, branch suppression helps inhibit the ability of the newly defrozen branch in the network to encode coarser-scale structures, thus ``encouraging'' it to encode the finer-scale structures in the new set of higher-resolution training images. Note that the inhibition or suppression is not absolute; the network weights in previously trained branches are still altered.

BSD-GAN progressively adds the depth (or layers) and breadth (or $\mathbf{z}$ sub-vectors) to the generator, and then start training with images of higher-resolution. During the process, branch suppression helps encourage the newly-added sub-vector to encode the finer-scale structures, as shown in Figure~\ref{fig:framework}. At each scale, a two-staged sub-procedure is used to avoid sudden shock to already well-trained, smaller-resolution layers: see Figure~\ref{fig:framework2}. Note that all already-added layers of the discriminator are trainable throughout the sub-procedure. More details about hyper-parameters are available in the supplementary material. 


The architecture of the generator is the same as shown in Figure~\ref{fig:framework}. For $256\times 256$ image generation, we use $5\mathbf{z}$ sub-vectors. The number of $\mathbf{z}$ sub-vectors is subject to change according to the resolution of output images. We use generator and discriminator networks that are mirror images of each other and always grow in synchrony, and use the standard non-saturated loss as in DCGAN~\cite{radford2015unsupervised} for training.


\subsection{Evaluation metrics for scale-disentanglement}

To examine how each dimension of the latent manifold space impacts appearance of the output images, we designed a metric to evaluate the variance of the output images when the latent vector is manipulated. The metric, which we refer to as {\em variance by scale\/} or {\em VBS}, denoted by $\mathcal V$, measures the variation of output images with respect to {\em any sub-vector\/}  $\mathbf{z'}$ of $\mathbf{z}$, at a specific scale, as reflected by a frequence interval $[f_1,f_2]$. That is, $\mathbf{z'}$ can correspond to a single dimension of $\mathbf{z}$ or to one of the designated sub-vectors ${\mathbf{z}^t}$, $t=0,\ldots,4$. Specifically,
\begin{equation}
\small
\label{eq:is1}
\begin{gathered}
{\mathcal V}_{f_1}^{f_2}(\mathbf{z'}) = \sum_{h,w,d}  \mathop{\mathbb{E}}_{\mathbf{c} \sim U(-1,1)} \sigma_{\mathbf{z'} \sim  U(-1,1), \overline{\mathbf{z'}} \leftarrow \boldsymbol{c}}, \big(\DFT_{f_1}^{f_2}(G({\mathbf{z}}))\big), \\
\text{and    } {\mathcal V}_{f_1}^{f_2}(\mathbf{z'}) = {\mathcal V}_{f_1}^{f_2}(\mathbf{z'}) / \mathop{\mathbb{E}}_{\mathbf{z'} \subseteq \mathbf{z}} {\mathcal V}_{f_1}^{f_2}(\mathbf{z'}),
\end{gathered}
\end{equation} 
where $\overline{\mathbf{z'}}$ is the set of dimensions of $\mathbf{z}$ excluding $\mathbf{z'}$. $\sigma_{\mathbf{z'} \sim  U(-1.0,1.0), \overline{\mathbf{z'}} \leftarrow \boldsymbol{c}} f(\mathbf{z})$ refers to the deviation of the value of $f(\mathbf{z})$ when $\mathbf{z'}$ follows the uniform distribution $U(-1.0,1.0)$ and $\overline{\mathbf{z'}}$ is fixed as a constant vector $\mathbf{c}$. $G(\mathbf{z})$ is the output image of the generator $G$ given $\mathbf{z}$. $h,w,d$ are the height, width, and depth of images (or layer activations). $\mathop{\mathbb{E}}(\boldsymbol{\cdot})$ is the expectation operator. In Eq.~(\ref{eq:is1}), $\DFT_{f_1}^{f_2}(\boldsymbol{\cdot})$ refers to the discrete Fourier transform of an image, and $(f_1,f_2)$ is a frequency range. In order to avoid the impact of image size, $\VBS$ is further normalized by dividing over their expected values. Intuitively, a larger value of $\VBS$ implies a greater impact of a manifold dimension (or a subset of manifold dimensions) on the output.

\subsection{Modified iGAN framework}

\begin{figure}[!t]
\begin{center}
\includegraphics[width=1.0\linewidth]{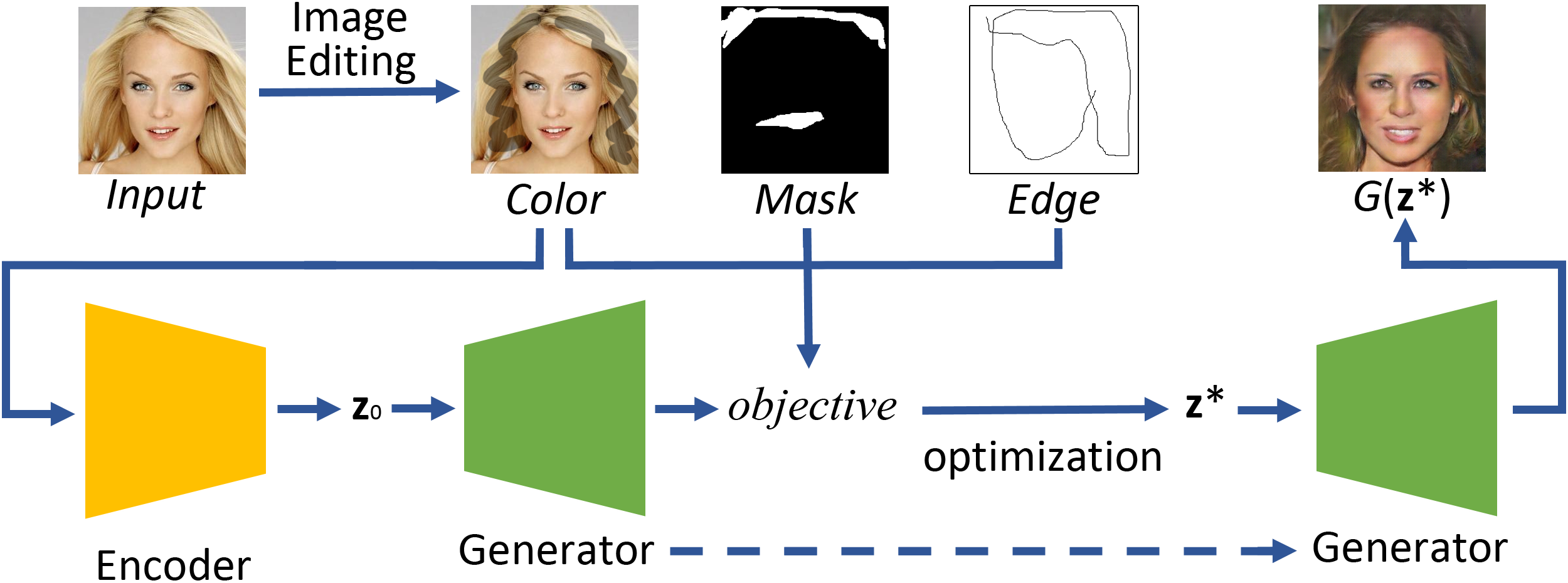}
\caption{ {\em Overview of modified iGAN workflow.} As in \cite{zhu2016generative}, the encoder that projects an image onto a manifold needs to be trained in advance.
A user-edited image 
is mapped to a latent code in the manifold space, which is assigned as the initial value ($\mathbf{z}_0$) of the latent vector. Then the latent vector $\textbf{z}$ is optimized to minimize the objective in Eq.~(\ref{eq:combined}).} \label{fig:overview}
\end{center}
\end{figure}

\begin{figure}[!t]
\begin{center}
\includegraphics[width=0.9\linewidth]{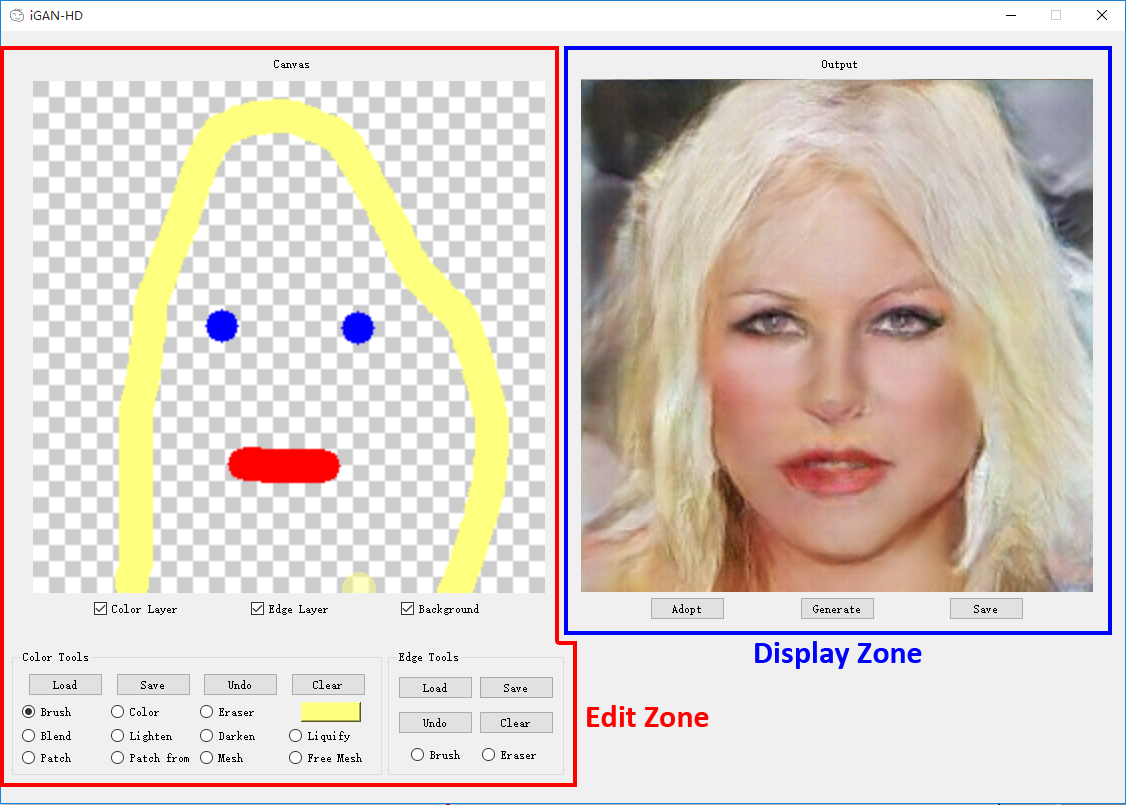}
\caption{ {\em GUI of the modified iGAN tool.} The main window includes an edit zone (left) and a display zone (right). The edit zone provides various tools and a canvas to help edit the color map and mask or produce the edge map.
The display zone shows the result generated by iGAN based on the edits. } \label{fig:ui-igan}
\end{center}
\end{figure}

The scale-disentangled image manifolds are expected to improve the interactive image editing results of iGAN \cite{zhu2016generative}. In the original iGAN~\cite{zhu2016generative} framework, a user makes interactive edits (eg., scribbles, warping) to an existing image. The edits are often casual and lead to various artifacts. The tool can automatically adjust the output image to keep all user edits as natural as possible. We conduct our experiment using a modified iGAN workflow, as shown in Figure~\ref{fig:overview}. 
Specifically, we abandoned ``edit transfer''~\cite{zhu2016generative} to improve performance for images with cluttered backgrounds, which is common in real applications. On the other hand, we enriched the available operations in the GUI, as shown in Figure \ref{fig:ui-igan}. We also removed the smoothness term from the original optimization objective to ensure more faithfulness to user inputs. Aside from these changes, the modified iGAN workflow is the same as the original iGAN implementation~\cite{zhu2016generative} and has the following objective consisting of a color term and an edge term:

\begin{equation}
\label{eq:combined}
\begin{gathered}
\mathbf{z}^*(\mathbf{C}, \mathbf{M},\mathbf{E}) = \arg\min_{\mathbf{z}} |\mathbf{C} - G(\mathbf{z})|\cdot \mathbf{M}/|\mathbf{M}|+ \\
\alpha |HOG(G(\mathbf{z})) - HOG(\mathbf{E})|,\\
\end{gathered}
\end{equation}
where $\mathbf{C}$ is the color map (edited image) and $\mathbf{M}$ is the mask. $\mathbf{M}(i,j)=1$ if pixel $(i,j)$ has color assigned and $\mathbf{M}(i,j)=0$ if $(i,j)$ has color erased. $|\mathbf{M}|$ denotes the sum of elements in $\mathbf{M}$. $\mathbf{E}$ is the edge map drawn with the edge tool and $\HOG(\cdot)$ is the differentiable $\HOG$ operator \cite{dalal2005histograms} which maps an image to a $\HOG$ descriptor. The parameter $\alpha$ is used to balance the two terms in the objective function. We set $\alpha=10$ throughout our experiment. Note that if the user does not provide an edge map, the edge term is disabled.

%% file: experiment.tex
\section{Experimental results}
\label{sect:exp}

We experimented BSD-GAN on three datasets: church\_outdoor from LSUN~\cite{yu15lsun}, CelebA\_HQ \cite{karras2017progressive}, and car. The original car dataset has 800 $\times$ 600 pixel resolution. To speed up the training, we used downsampled versions of CelebA\_HQ (256 $\times$ 256 and 512 $\times$ 512) and car (400 $\times$ 300). We trained each model on a single GTX TITAN XP GPU, for 45 epochs at each scale, using Adam optimizer with a fixed learning rate of 0.0002. The codes are implemented with tensorflow v1.4.0, CUDA v8.0 and cuDNN v5.1.

\subsection{Evaluation of scale disentanglement}

\begin{figure}[!t]
\begin{center}
\includegraphics[width=.9\linewidth]{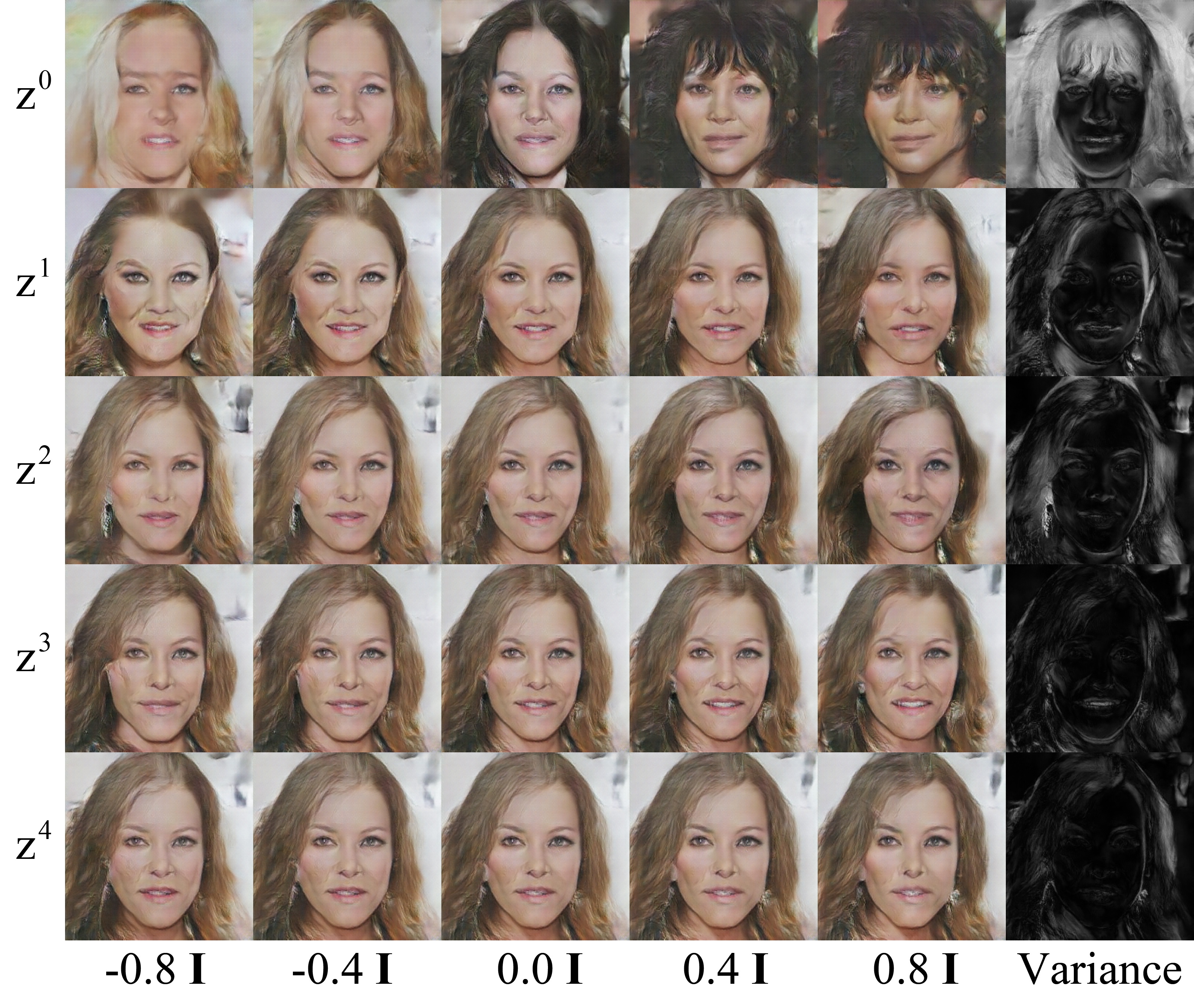}
\vspace{-10pt}

\caption{ {\em Generated images for CelebA\_HQ dataset by varying individual sub-vectors.\/} We first initialize ${\mathbf{z}}$ randomly, and then replace one of the sub-vectors ${\mathbf{z}^t}$, $t=0,\ldots,4$, by $p{\bf I}$, where $p=-0.8,-0.4,0,0.4,$ or $0.8$ and ${\bf I}$ is the all-one vector of length $|{\mathbf{z}^t}|$, while holding all the other sub-vectors
fixed. Columns $1$ to $5$ show images generated by BSD-GAN and the last column shows a variance image for the five generated images on the left, where lightness reflects pixel variance. From top to bottom, changing ${\mathbf{z}^t}$ leads to smaller and smaller image variations, as reflected by intensity drop in the variance images. Sub-vector ${\mathbf{z}^0}$ dominates the overall color, ${\mathbf{z}^1}$ controls some facial features, while the rest bring minor changes near ear, mouth, and hair.}
\label{fig:face256-1}
\end{center}
\vspace{-10pt}

\end{figure}

\begin{figure}[!t]
\begin{center}
\includegraphics[width=\linewidth]{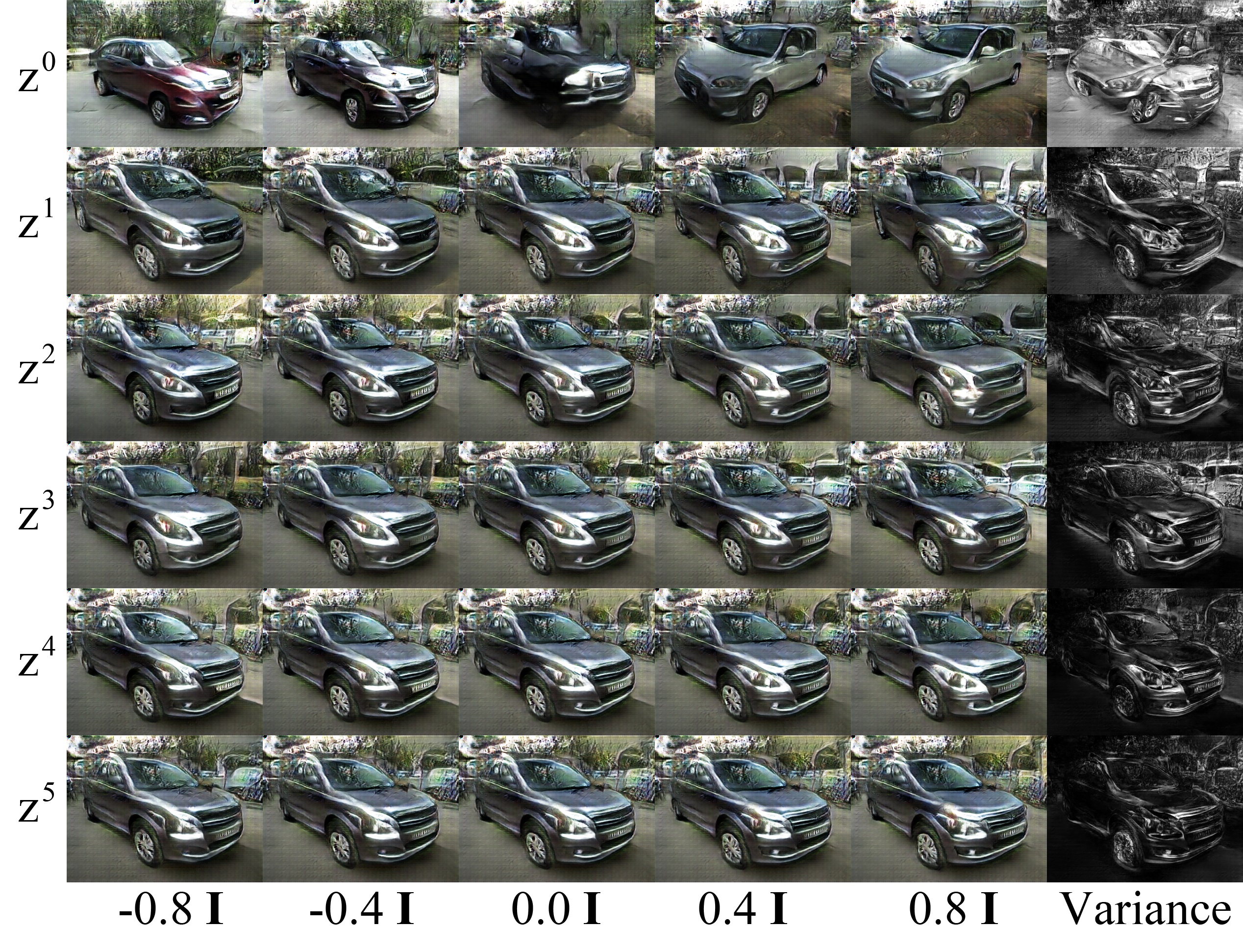}
\vspace{-10pt}

\caption{{\em Generated images for car dataset by varying individual sub-vectors.\/} The output setting is the same as in Figure~\ref{fig:face256-1}. The first row basically reflects the property of the $car$ dataset: it contains left-views and right-views, but no front-views. GAN was not able to learn a smooth interpolation between the two views, resulting in a messy image in the middle of the first row. However, once ${\mathbf{z}^0}$ is fixated with a view, the other sub-vectors can generate smooth interpolations. ${\mathbf{z}^1}$ appears to alter the view angles slightly, ${\mathbf{z}^2}$ impacts the front parts of the car, while changing the other sub-vectors influences more minor details.} 
\label{fig:car}
\end{center}
\vspace{-10pt}

\end{figure}

\begin{figure}[!t]
\begin{center}
\includegraphics[width=.9\linewidth]{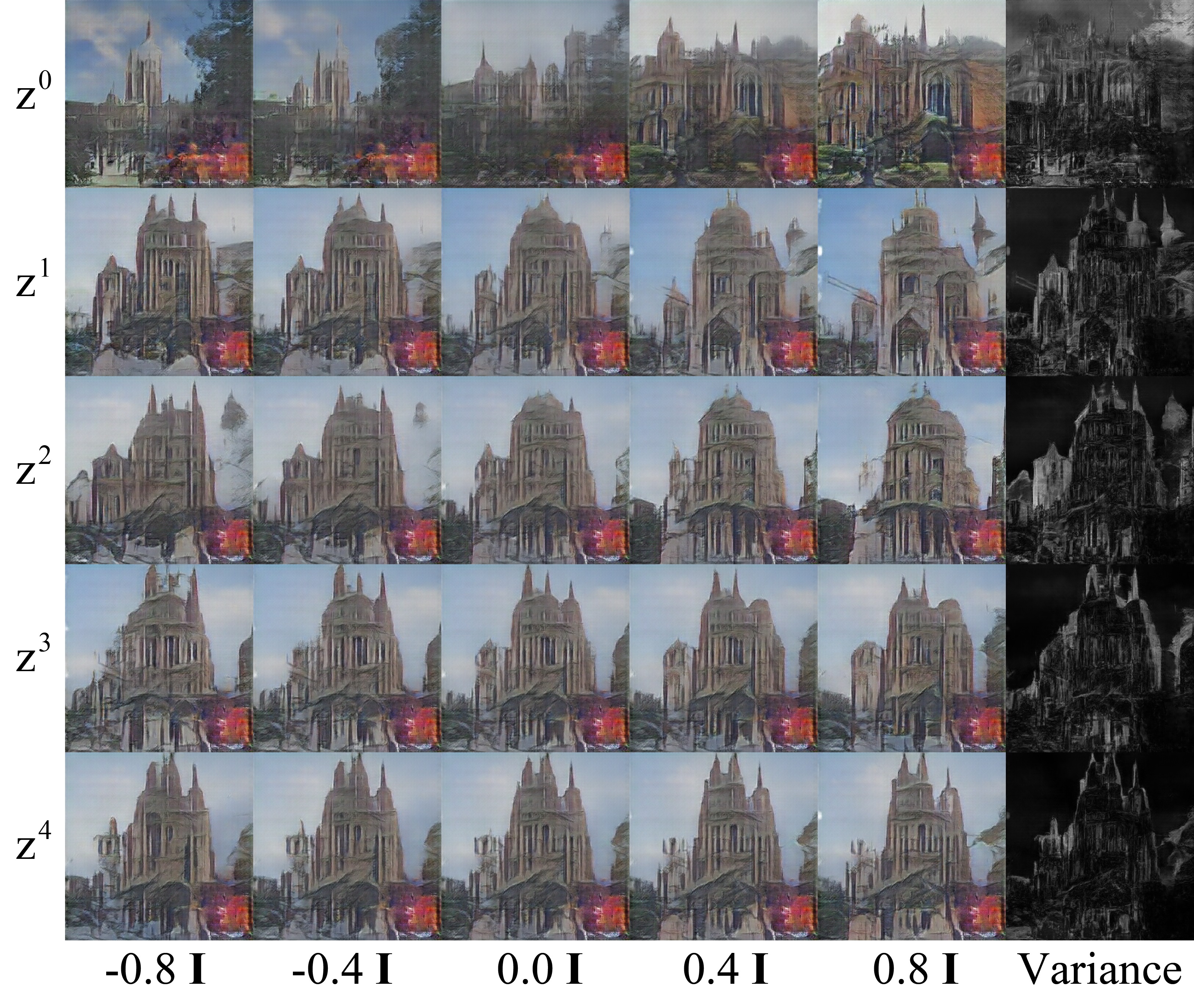}
\vspace{-10pt}

\caption{ {\em Generated images for church\_outdoor dataset by varying individual sub-vectors.\/} The output settings is the same as in Figure~\ref{fig:face256-1}. Similarly, as more clearly reflected in the variance images, the sub-vectors ${\mathbf{z}^t}$, with $t$ from 0 to 4, appear to control higher-level to finer image details.} 
\label{fig:lsun}
\end{center}
\vspace{-10pt}

\end{figure}

\paragraph{Qualitative evaluation.}
We wish to show how each designated sub-vector affects images generated via BSD-GAN training. To this end, we vary the values of each sub-vector while holding the other sub-vectors fixed, as shown in Figures~\ref{fig:face256-1}, \ref{fig:car}, and \ref{fig:lsun}. From visual examination and as detailed in the figure captions, we can observe that $\mathbf{z}^0$ affects the output images most significantly as it mainly controls large-scale structures. In contrast, the effects of $\mathbf{z}^4$ or $\mathbf{z}^5$ are closely tied to finer-scale details. This suggests that scale disentanglement by branching the training to progressively activated sub-vectors has been achieved. That being said, it is also clear that we do not have direct control over feature localization or semantic image manipulation.

\begin{figure}[!t]
\begin{center}
\includegraphics[width=\linewidth]{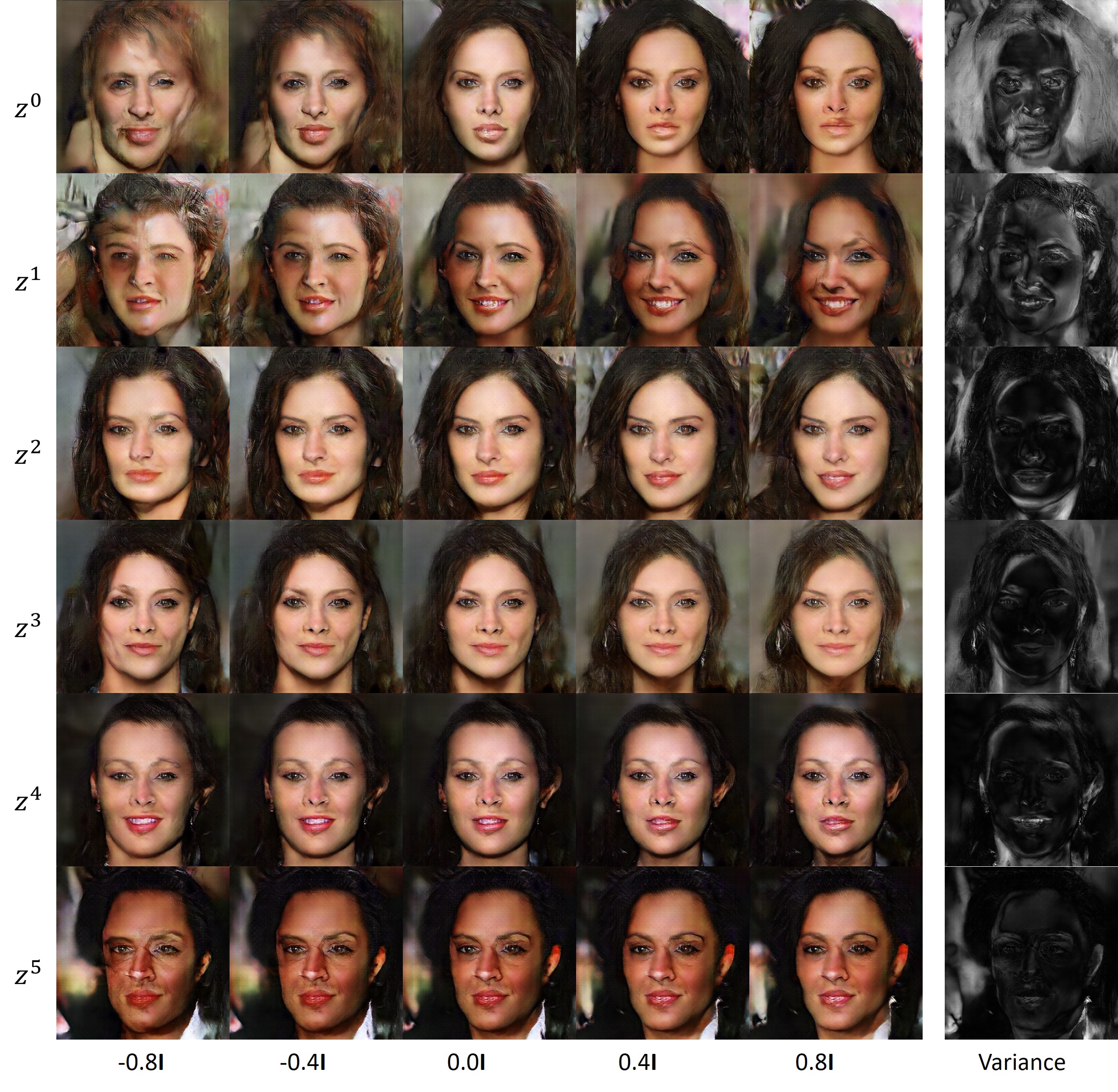}
\vspace{-10pt}

\caption{{\em Effects on generated images for CelebA\_HQ 512 $\times$ 512 dataset by varying individual sub-vectors.\/} The output setting is similar as in Figure~\ref{fig:face256-1} except that there is one more sub-vector $\mathbf{z}^5$ and each row is generated independently. Similar to Figure~\ref{fig:face256-1}, sub-vector ${\mathbf{z}^0}$ dominates the overall color, ${\mathbf{z}^1}$ controls some facial features and hair features, while the rest bring minor changes near ear, mouth, and hair.} \label{fig:face512-branch}
\end{center}
\vspace{-10pt}

\end{figure}

Figure \ref{fig:face512-branch} shows qualitative results of BSD-GAN on higher-resolution version (512$\times$512) of CelebA\_HQ dataset.

\noindent
\noindent

\begin{figure}[!t]
\begin{center}
\includegraphics[width=\linewidth]{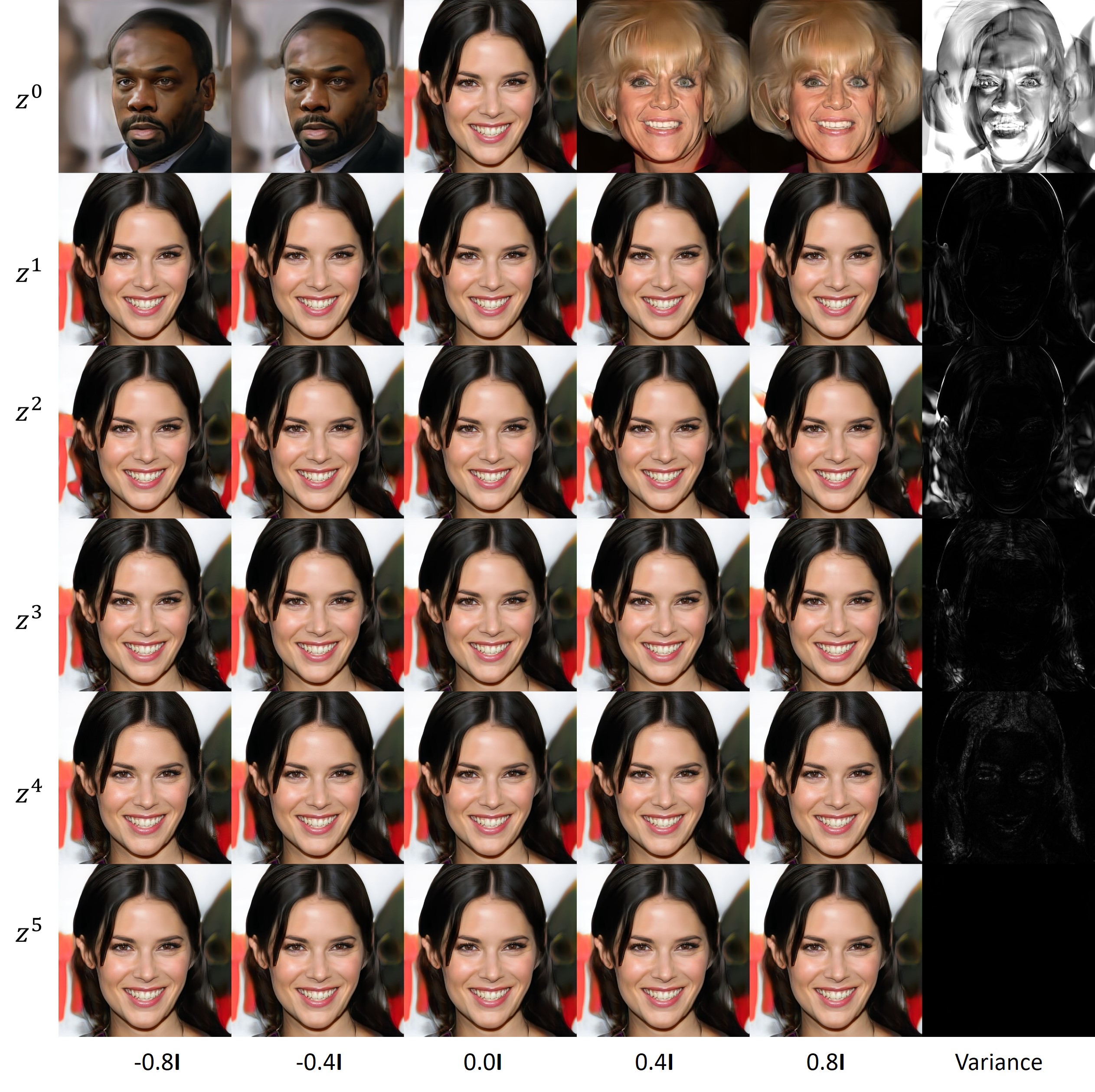}
\vspace{-10pt}

\caption{ {\em Generated images for CelebA\_HQ dataset by varying individual latent sub-vectors of StyleGAN.\/} The output setting is similar as in Figure~\ref{fig:face256-1} except that there is one more sub-vector $\mathbf{z}^5$. As reflected by intensity drop in the variance images, sub-vector ${\mathbf{z}^0}$ dominates the overall color, while the rest bring minor changes near ear, mouth, and hair.}
\label{fig:face256-style}
\end{center}
\vspace{-10pt}

\end{figure}

For comparison, we here show qualitative evaluation results of StyleGAN~\cite{karras2019style}, a concurrent work carried out by Karras et al. that targeted on the same scale-disentangle purpose: see Figure \ref{fig:face256-style}. As the latent codes of StyleGAN consist of 1 style vector and 18 noise chunks, we only select a subset of them for illustration convenience. We set the style vector as $\mathbf{z}^0$ and 5 noise chunks from increasingly smaller scale levels as $\mathbf{z}^t, t=1,\ldots,5$. We use the official pretrained model to generate $1024\times1024$ images and observe how generated images are controlled by each sub-vector. As shown in Figure~\ref{fig:face256-style}, the style vector dominants controls over the overall color as well as large- and middle-scale structures, while other noise chunks have minor impacts on finest details. Specifically, $\mathbf{z}^1$ has less impact than $\mathbf{z}^2$ while $\mathbf{z}^5$ appears to make no noticeable difference, implying that StyleGAN is inferior to BSD-GAN in terms of scale-disentanglement, 

\begin{figure}[!t]
\begin{center}
\includegraphics[width=.9\linewidth]{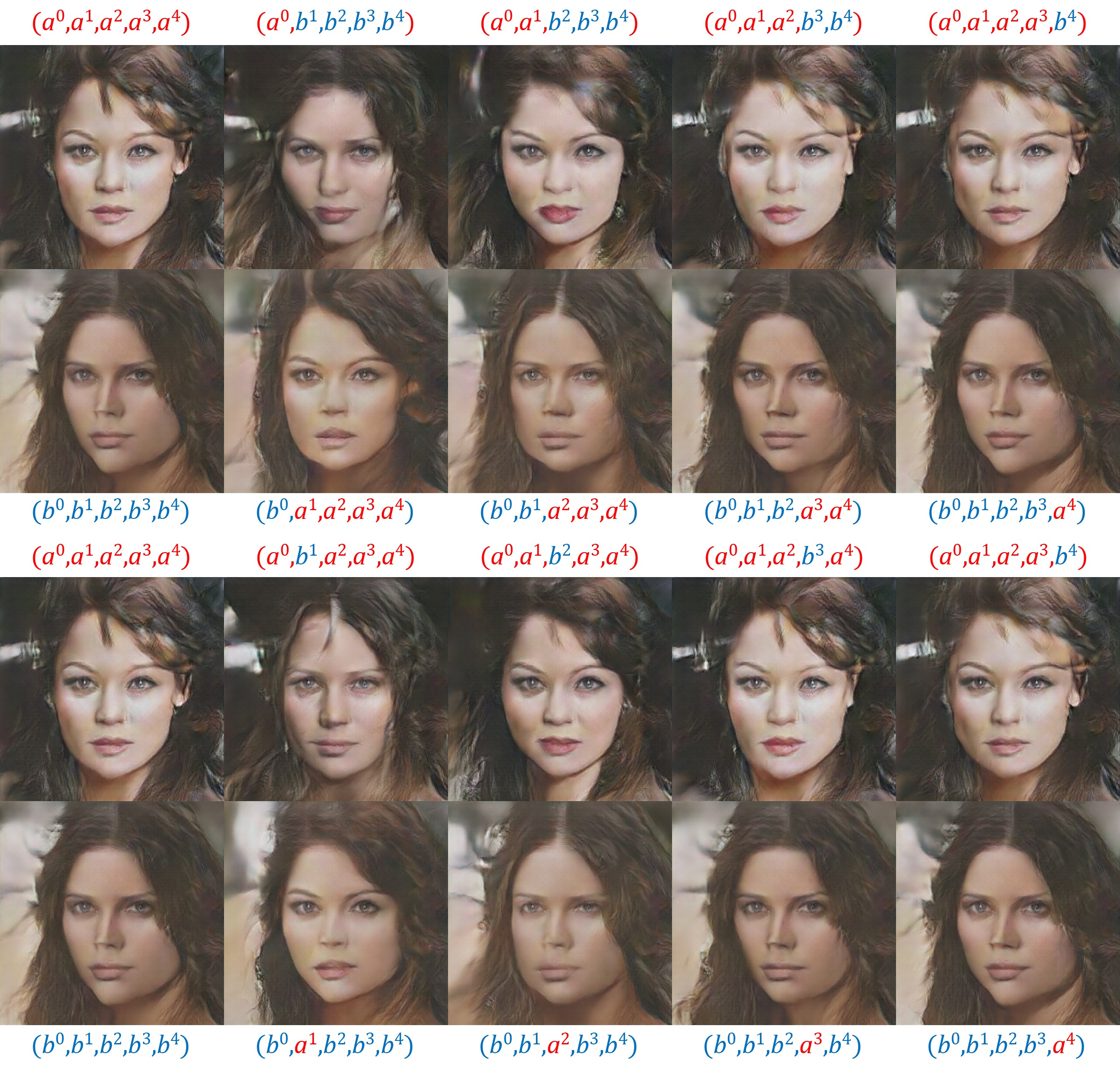}
\vspace{-10pt}

\caption{{\em Cross-scale image fusion.\/} The notations and synthesis setup are the same as in Figure~\ref{fig:teaser}. From the first three columns, the sub-vector $\mathbf{x}^1$ mainly controls the facial features while $\mathbf{x}^2$ controls the face shape. By swapping $\mathbf{x}^1$ and $\mathbf{x}^2$, a face swap may be achieved. $\mathbf{x}^3$ and $\mathbf{x}^4$ have less significant impacts, such as lighting, shading, and minor changes in hair and ear.}
\label{fig:fusion}
\end{center}
\vspace{-10pt}

\end{figure}

\begin{figure*}
\centering
\includegraphics[width=.8\linewidth]{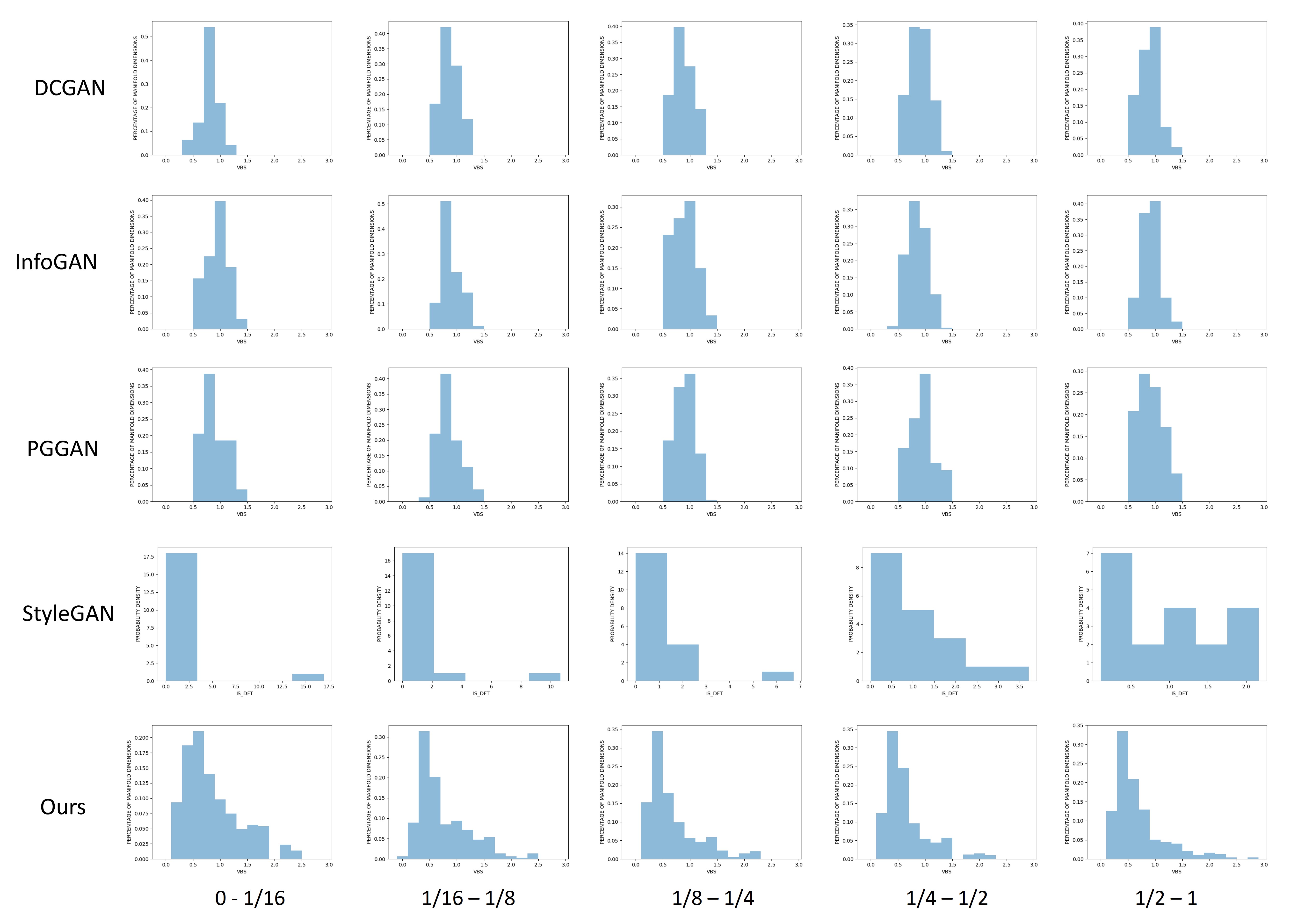}
\vspace{-10pt}
\caption{ {\em Distributions of $\VBS$ over specific frequency ranges or image scales.\/} The $\VBS$ of DCGAN~\cite{radford2015unsupervised}, PGGAN \cite{karras2017progressive} and InfoGAN \cite{chen2016infogan} predominantly falls into the interval $(0.5, 1.5)$, whereas the $\VBS$ of BSD-GAN and StyleGAN\cite{karras2019style} span over a wider range ($(0.1, 2.5)$ for BSD-GAN, and $(0.0, 10.5)$ for StyleGAN respectively).} 
\label{fig:VBS_histo} 
\vspace{-10pt}

\end{figure*}

\begin{figure}
\centering
\includegraphics[width=\linewidth]{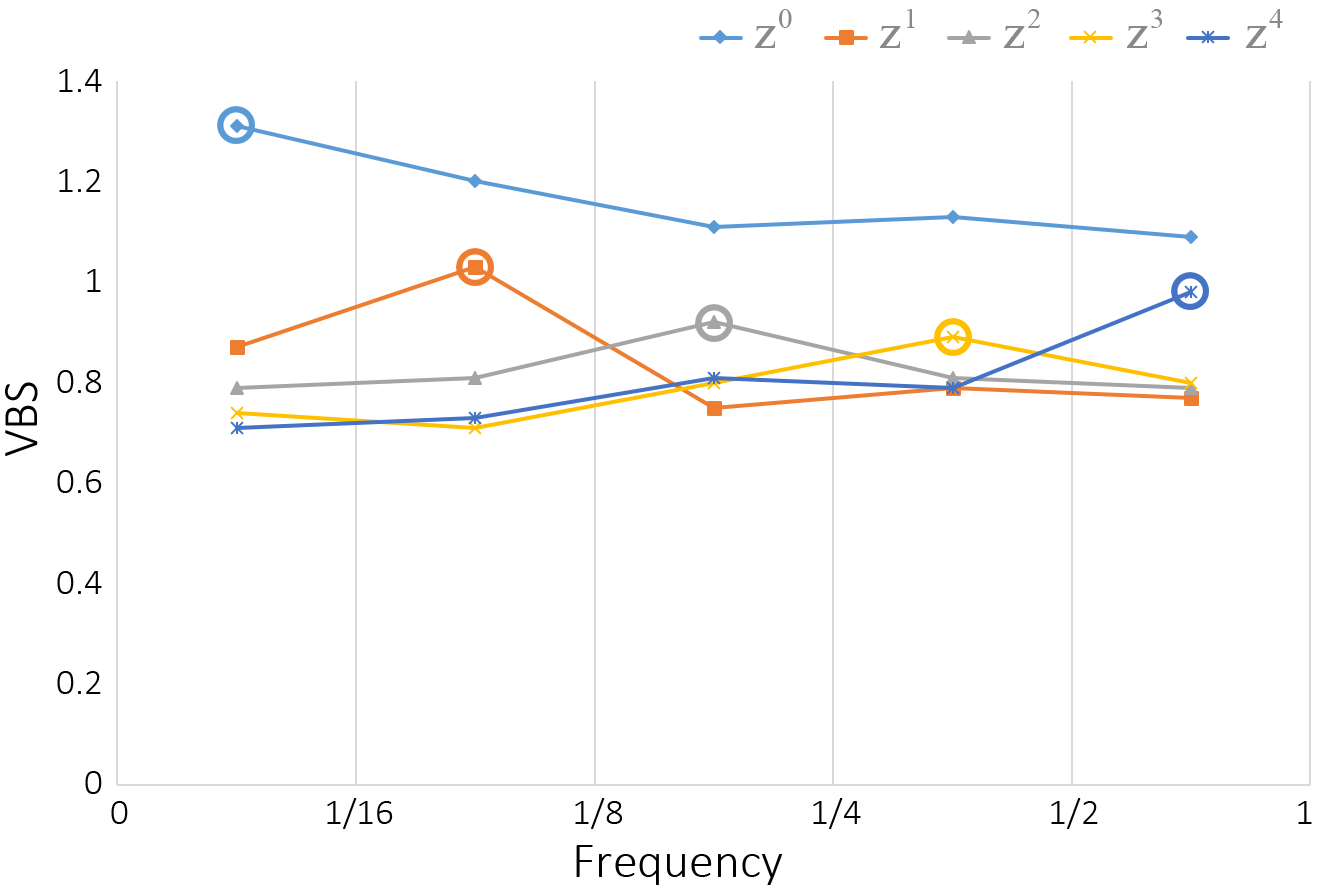}
\vspace{-10pt}

\caption{{\em Plot of $\VBS$ values for sub-vectors $\mathbf{z}^0, \mathbf{z}^1, \ldots, \mathbf{z}^4$ of BSD-GAN against frequencies.\/} Peak $\VBS$ values indicate maximal impact. For example, sub-vector $\mathbf{z}^0$ exhibits higher impact over the lowest frequency range, which corresponds to larger image scales.} 
\label{fig:VBS_peak}
\vspace{-10pt}

\end{figure}

\begin{figure}
\centering
\includegraphics[width=\linewidth]{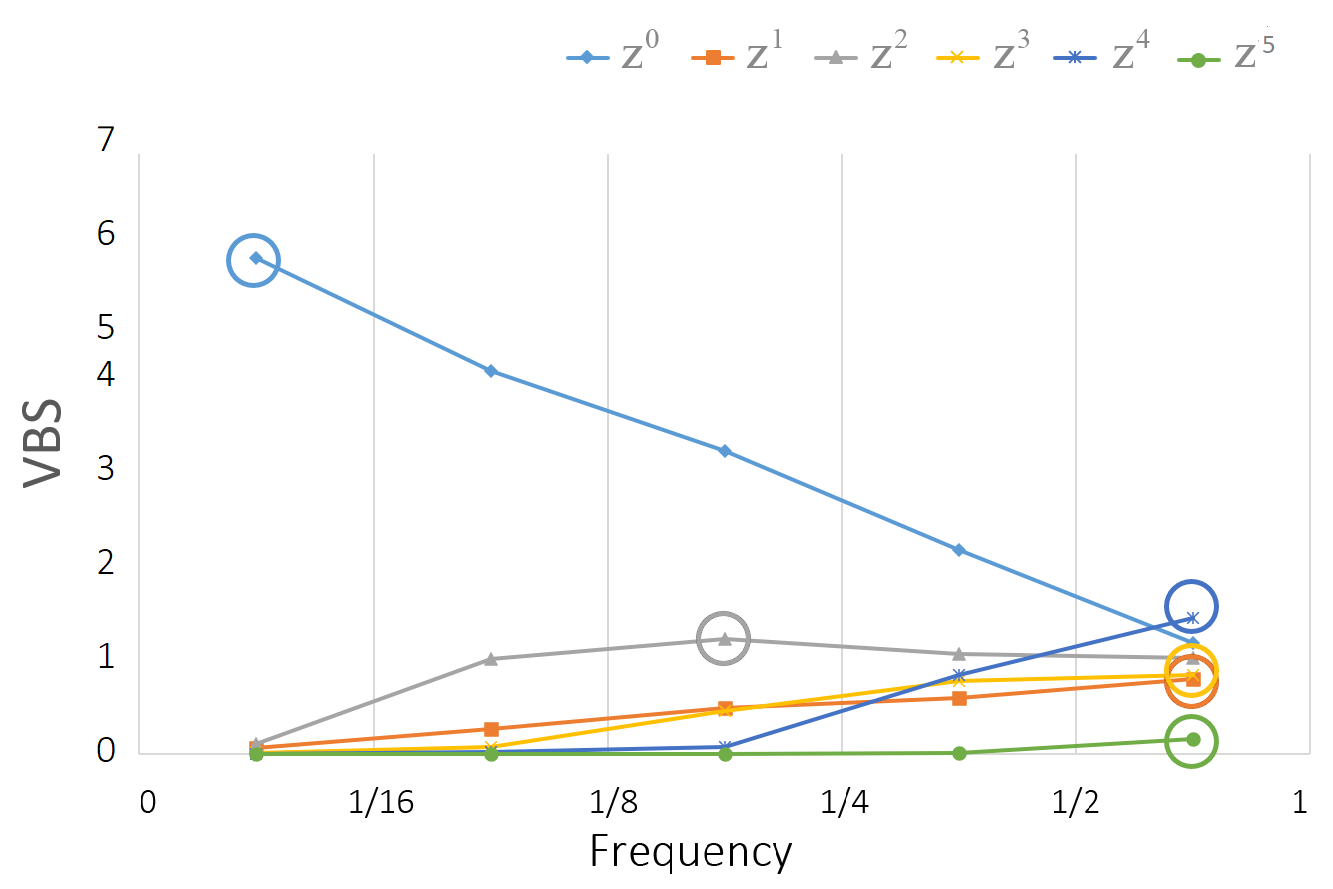}
\vspace{-10pt}

\caption{{\em Plot of $\VBS$ values for latent sub-vectors $\mathbf{z}^0, \mathbf{z}^1, \ldots, \mathbf{z}^5$ of StyleGAN against frequencies.\/} Peak $\VBS$ values indicate maximal impact. As shown, most noise chunks see their greatest impact on the highest frequency range, implying that they only encode the small-scale image structures.} 
\vspace{-10pt}

\label{fig:VBS_peak_stylegan}
\end{figure}

\paragraph{Quantitative evaluation.}

We use the proposed $\VBS$ metrics to quantitatively evaluate the scale-disentanglement. To examine the distributions of $\VBS$ at specific scales, we split the frequency domain into five ranges: $(0,1/16)$, $(1/16, 1/8)$, $(1/8, 1/4)$, $(1/4, 1/2)$, and $(1/2, 1)$, which roughly correspond to increasingly fine image scales. We then visualize the $\VBS$ distributions for various GANs using histogram plots, as shown in Figure~\ref{fig:VBS_histo}. To produce the histograms, we randomly sampled $10\mathbf{z}$ vectors to generate 10 images using the trained model for the CelebA\_HQ dataset. For each $\mathbf{z}$ vector, which is of dimension 150, and for each dimension, we vary it while keeping all the other dimensions fixed. This results in a set of varied images, from which we compute the $\VBS$ value for the selected dimension. Overall, we collect $10 \times 150 = 1,500$ $\VBS$ values to form the histograms, for each GAN option and for each frequency range.

As shown in Figure~\ref{fig:VBS_histo}, the $\VBS$ values of BSD-GAN and StyleGAN\cite{karras2019style} exhibit a much greater variance (i.e., wider histogram) than those of traditional GANs, such as DCGAN~\cite{radford2015unsupervised}, PGGAN \cite{karras2017progressive}, and InfoGAN \cite{chen2016infogan}. Specifically, the $\VBS$ values of these traditional GANs at all scale levels mainly fall into the range of $[0.5, 1.5]$, implying that the corresponding image representations are more scale-entangled, in comparison to BSD-GAN, whose $\VBS$ values vary over a larger interval $[0.1, 2.5]$. Similar to BSD-GAN, $\VBS$ values of StyleGAN also span over a wide range of $(0.0, 17.0)$.

Finally, to examine whether the image representations learned by BSD-GAN are disentangled by the designated sub-vectors $\mathbf{z}^t, t = 0,\ldots,4$, we show how the $\VBS$ of each $\mathbf{z}^t$ varies against frequencies in Figure~\ref{fig:VBS_peak}. These $\VBS$ values were obtained in the same way as for the histogram plots in Figure~\ref{fig:VBS_histo}. As we can observe, $\mathcal V(\mathbf{z}^0)$ sees its peak value in the $(0, 1/16)$ range, implying that $\textbf{z}^0$ mainly controls larger-scale structures of the generated images. The remaining sub-vectors $\mathbf{z}^1$, $\mathbf{z}^2$...$\mathbf{z}^4$ show their peaks at frequency intervals which reflect their respective controls over increasingly finer image features or structures.

Similarly, we show  the $\VBS$ of each latent vector $\mathbf{z}^t, t = 0,\ldots,5$ of StyleGAN against frequencies in Figure~\ref{fig:VBS_peak_stylegan}. As the latent codes of StyleGAN consist of 1 style vector and 18 noise chunks, we only select a subset of them for illustration convenience. We set the style vector as $\mathbf{z}^0$ and 5 noise chunks from different scales as $\mathbf{z}^t, t=1,\ldots,5$ (the same as described in Figure \ref{fig:face256-style}). As shown in Figure~\ref{fig:VBS_peak_stylegan}, the style vector dominants controls over coarse image structures, while other noise chunks have more impacts on finer scale structures. Unlike BSD-GAN which disentangles image structure into 5 scales, StyleGAN can only disentangle image structures into two or three scales.

\subsection{Ablation study}
\label{subsec:ablation}

\begin{figure*}[!t]
\begin{center}
\includegraphics[width=.9\linewidth]{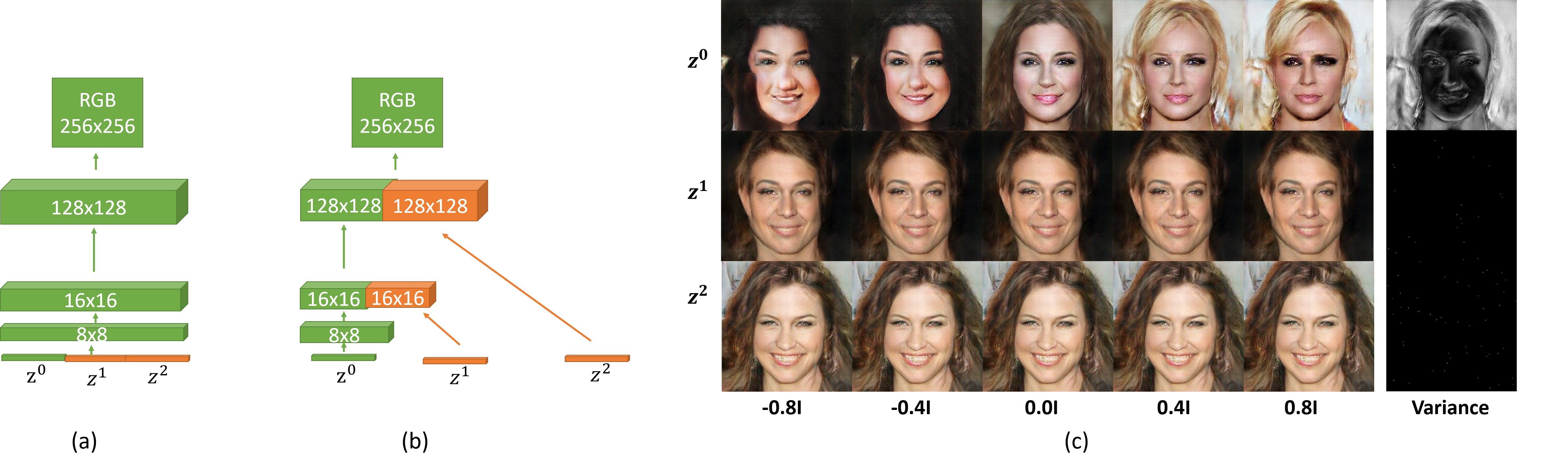}
\vspace{-10pt}

\caption{{\em Ablation studies of design of Generator}. (a) all sub-vectors are fed through the input layer. (b) different sub-vectors are provided to respective hidden layer. We use the same training procedure for both variations. For convenience of presentation, we only visualize three sub-vectors ($\mathbf{z}^t, t \in \{1,2,3\}$). (a) is what we used in our method and the results are demonstrated in Figure \ref{fig:face256-1}. The results of (b) are visualized in (c), which show that adjusting the value of the newly-joined sub-vectors (($\mathbf{z}^1$, $\mathbf{z}^2$) do not change the generated images at all.} \label{fig:ablation}
\end{center}
\vspace{-10pt}

\end{figure*}

As shown in Figure \ref{fig:framework} and \ref{fig:framework2}, in the design of generator we split the latent variable into scale-specific pieces and feed all scales through the input layer of the generator. A more obvious way is to provide the different latent chunks at the respective hidden layer. We experimented and compared both variations: see Figure \ref{fig:ablation}. We found that the latter variation causes another form of ``branch suppression''. Unlike the one discussed in Section \ref{sect:framework}, the new form of ``branch suppression'' comes that newly-added branches do not encode any information at all, no matter how long it is trained. More experiments show that it does happen whenever new branches are provided through concatenation or addition, whenever the latent chunks are repeated, convolved or linearly mapped before joining the hidden layer. As shown in Figure \ref{fig:ablation}, the newly-added latent chunks ($\mathbf{z}^1$, $\mathbf{z}^2$) do not contribute any variation to the generated images, implying they do not encode any information.

\noindent

\subsection{Applications}

\begin{figure}[!t]
\begin{center}
\includegraphics[width=\linewidth]{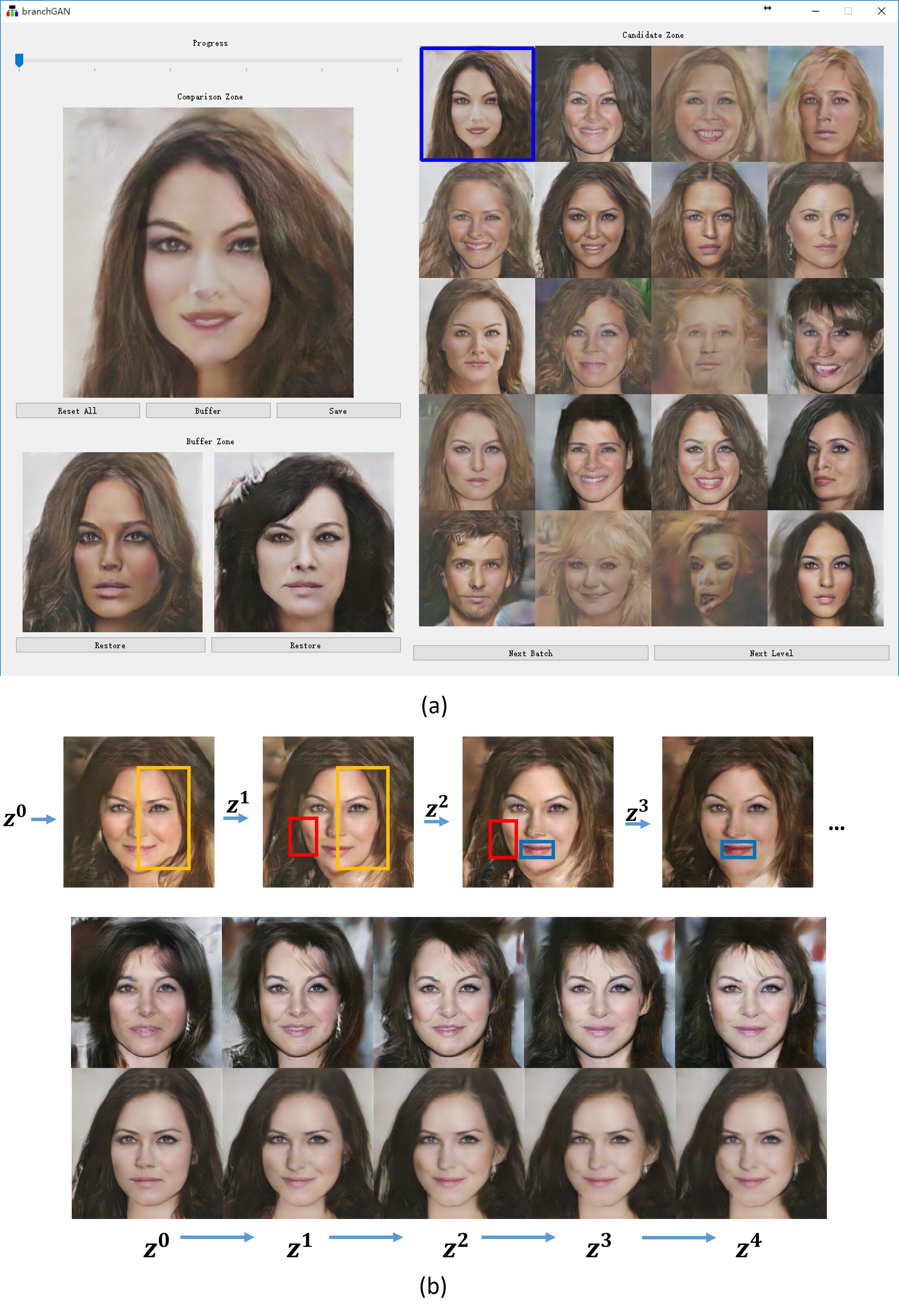}
\vspace{-10pt}

\caption{{\em Coarse-to-fine image synthesis.\/} (a): GUI of the application. (b): Sequences of images selected in a coarse-to-fine manner. Bounding boxes with the same color highlight the changes made by the user at each step: yellow box $\rightarrow$ thinner face, red box $\rightarrow$ less dimple, and blue box $\rightarrow$ red lip.} 
\label{fig:app1}
\end{center}
\vspace{-10pt}

\end{figure}

\paragraph{Coarse-to-fine image synthesis}
We show that the scale disentanglement afforded by BSD-GAN facilitates coarse-to-fine image synthesis. To this end, we developed a new interactive application; see Figure~\ref{fig:app1} (a). A user can select best-matching faces from randomly-generated ones displayed on the right panel. At the coarsest scale, the images are mapped from different $\mathbf{z}^0$ values with other sub-vectors set to zero. If the user is satisfied with a coarse-level image, then he/she can select it and move on to the next scale. Then, the value of $\mathbf{z}_0$ will be fixed and images mapped from different $\mathbf{z}^1$ values will be displayed for selection. As a result, the user can progressively improve the appearance of a synthesized face, as shown in Figure~\ref{fig:app1} (b). 


\paragraph{Cross-scale image fusion}
Scale disentanglement facilitates another new application: {\em cross-scale image fusion}, where latent codes representing different scales are joined to create hybrid images. Figures~\ref{fig:teaser} and~\ref{fig:fusion} show some examples of such image fusion, which are synthesized by integrating coarse-scale features of one image with fine-scale features of another. Through swapping sub-vectors representing different scales, our approach can achieve feature exchanges at both coarse-level (e.g. face swap) and fine-level (e.g. expression and face shape transfer). 


\paragraph{Improving interactive image editing (iGAN)}

We show how BSD-GAN can improve the performance of iGAN~\cite{zhu2016generative}. As mentioned in the method section, we have modified the objective to better handle the test datasets and enriched the functionality. In the original paper, DCGAN~\cite{radford2015unsupervised} was adopted. We now compare different choices of GANs as replacement for DCGAN, including PGGAN~\cite{karras2017progressive}, InfoGAN~\cite{chen2016infogan}, StyleGAN ~\cite{karras2019style}, and BSD-GAN. To examine which GAN manifold serves iGAN better, we used the minimum optimization value of the objective in Eq.~(\ref{eq:combined}) as the metric. The minimum optimization loss indicates to what extent the output image fits user inputs. Smaller objective value represents greater effectiveness of the GAN manifold in avoiding local-minima traps. We compared the minimum optimization loss of DCGAN~\cite{radford2015unsupervised}, PGGAN \cite{karras2017progressive}, InfoGAN \cite{chen2016infogan}, StyleGAN ~\cite{karras2019style}, and our BSD-GAN, given the same set of user inputs. To simplify the comparison, we used 60 images from the training dataset as the user inputs. Table \ref{table:error} shows that BSD-GAN has the lowest minimum objective value among all GANs, implying that the scale-disentangled representations learned by BSD-GAN can better fit user inputs. The reason that StyleGAN performs worse than ours probably lies on its failure to disentangle image structures into multiple scales.


\begin{table}[!t]
\begin{center}
\begin{tabular}{c||c|c|c|c}
Dataset   & \multicolumn{2}{c|}{CelebA\_HQ} &  car  & church\_outdoor \\
\hline
Image size & $256^2$ & $512^2$ & $400 \times 300$  & $256^2$  \\
		\hline	
		\hline
DCGAN  & 0.24      &   0.26 &  0.29 & 0.27\\
\hline
PGGAN  &     0.18   &  0.17  &  0.19 & 0.22\\
\hline
InfoGAN &  0.23      &   0.25 &  0.29 & 0.24    \\
\hline
StyleGAN &  0.18      &   0.16 &  -- & -- \\
\hline
Ours &  \textbf{0.15}        & \textbf{0.14}   &   \textbf{0.17}  & \textbf{0.18} \\
\end{tabular}
\caption{Average minimum optimization loss of iGAN using different GANs. DCGAN was used by the original iGAN.} 
\label{table:error}
\end{center}
\vspace{-10pt}

\end{table}

\begin{figure}[!t]
\begin{center}
\includegraphics[width=\linewidth]{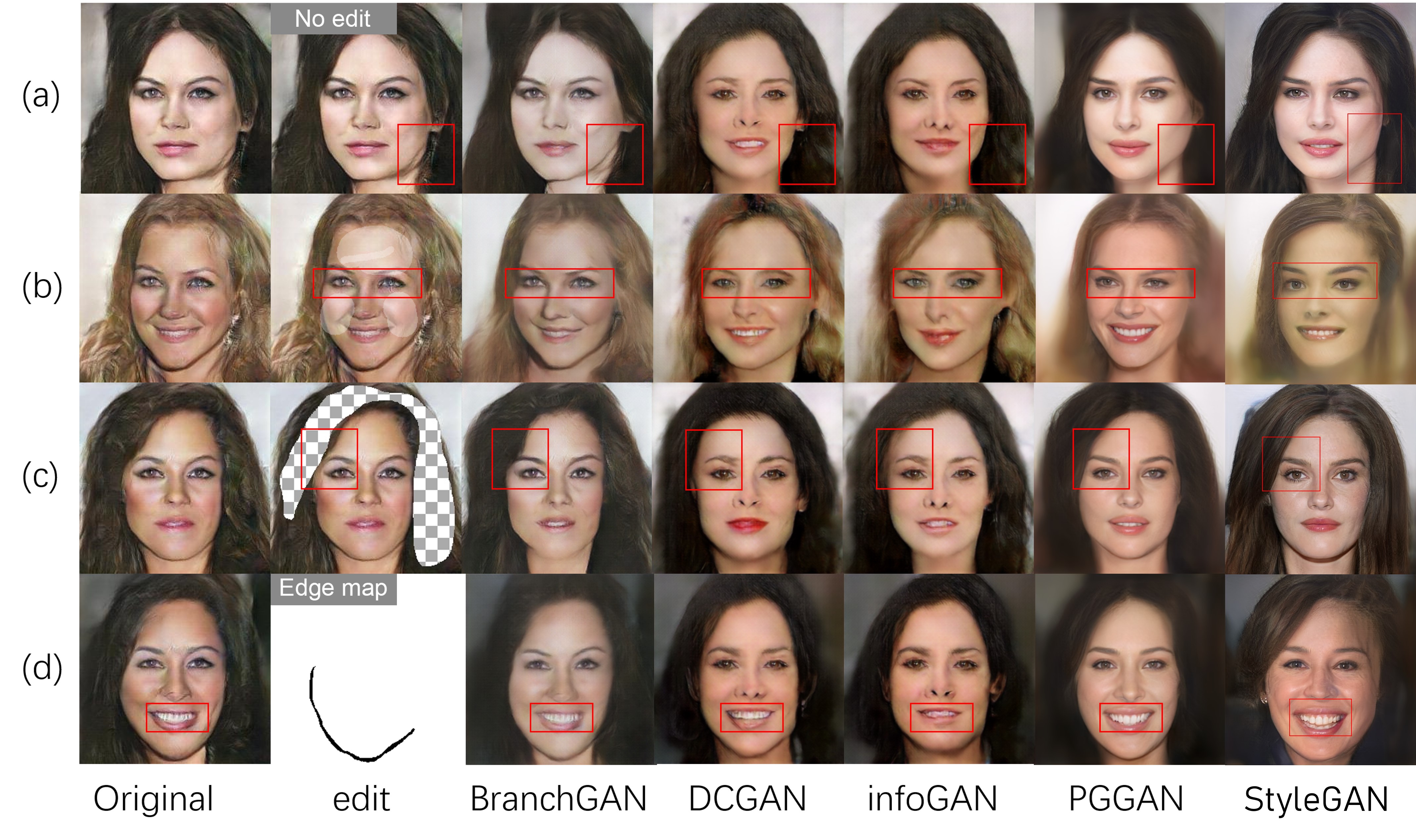}
\vspace{-10pt}

\caption{{\em Comparison of image editing results using different GAN manifolds as the latent space.\/} Images are edited with various user inputs: (a) no edit, (b) face lightened, (c) hair erased, (d) adding edge map. Notably, none of the results perfectly fit user inputs due to mode collapse. The red boxes highlight the details where some results fail to fit.} \label{fig:edit-face}
\end{center}
\vspace{-10pt}

\end{figure}

Figure~\ref{fig:edit-face} shows a few image editing examples using different GAN representations as the latent space of iGAN. The visual comparison between BSD-GAN and other GANs indicates that BSD-GAN generally performs better in fitting both coarse-level structures and fine-scale features.

\begin{figure*}[!t]
\begin{center}
\includegraphics[width=0.9\linewidth]{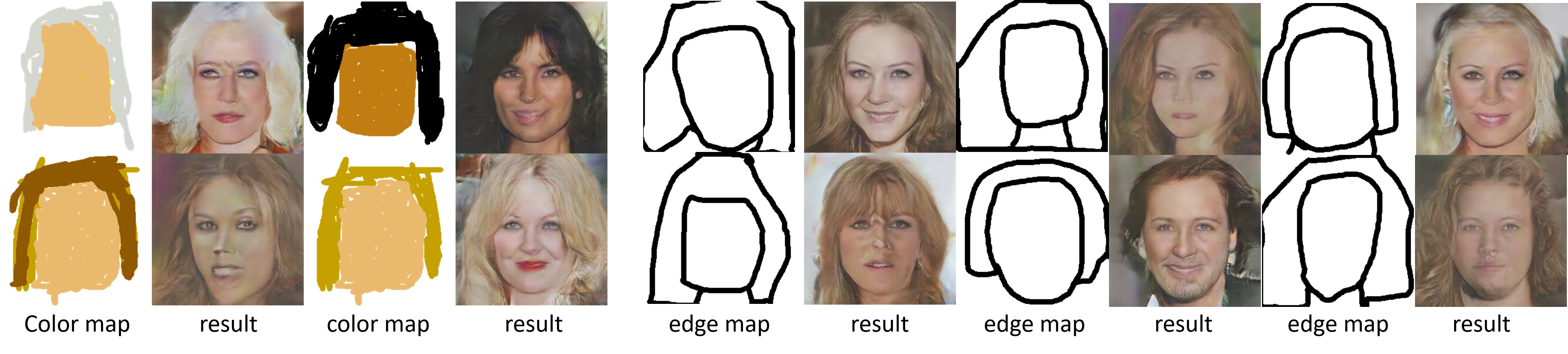}
\vspace{-10pt}

\caption{The edge maps and color maps drawn by users and the corresponding image generation results with improved iGAN.} \label{fig:face256}
\end{center}
\vspace{-10pt}

\end{figure*}

\begin{figure}[!t]
\begin{center}
\includegraphics[width=\linewidth]{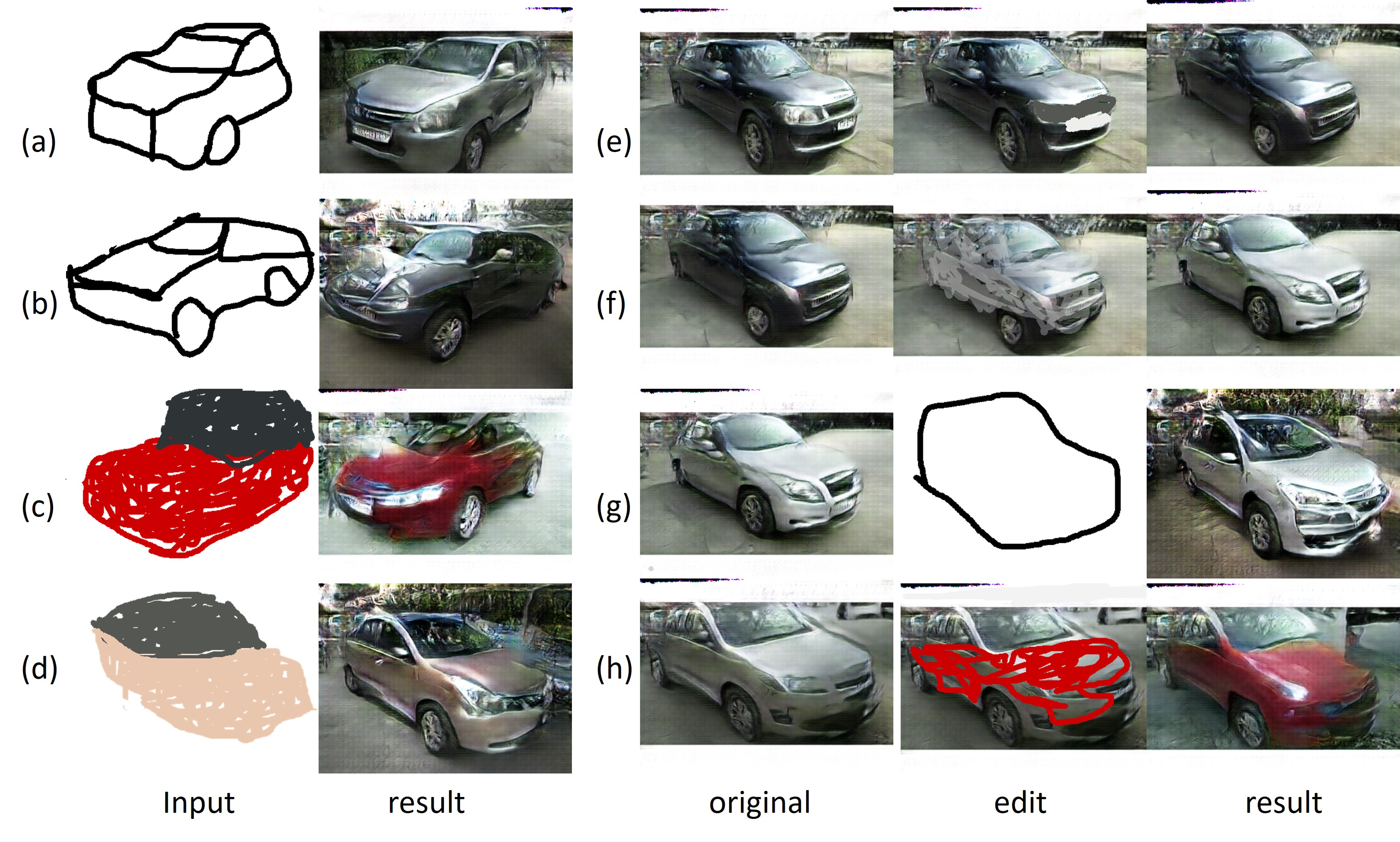}
\vspace{-10pt}

\caption{Car image generation and editing results with improved iGAN: (a-b) results based on edge maps; (c-d) results based on masked color maps; (e-h) manipulation of existing images, including erasing license plate (e), changing body color (f \& h), and adding extra edge map (g).}
\label{fig:car-igan}
\end{center}
\vspace{-10pt}

\end{figure}

\begin{figure}[!t]
\begin{center}
\includegraphics[width=\linewidth]{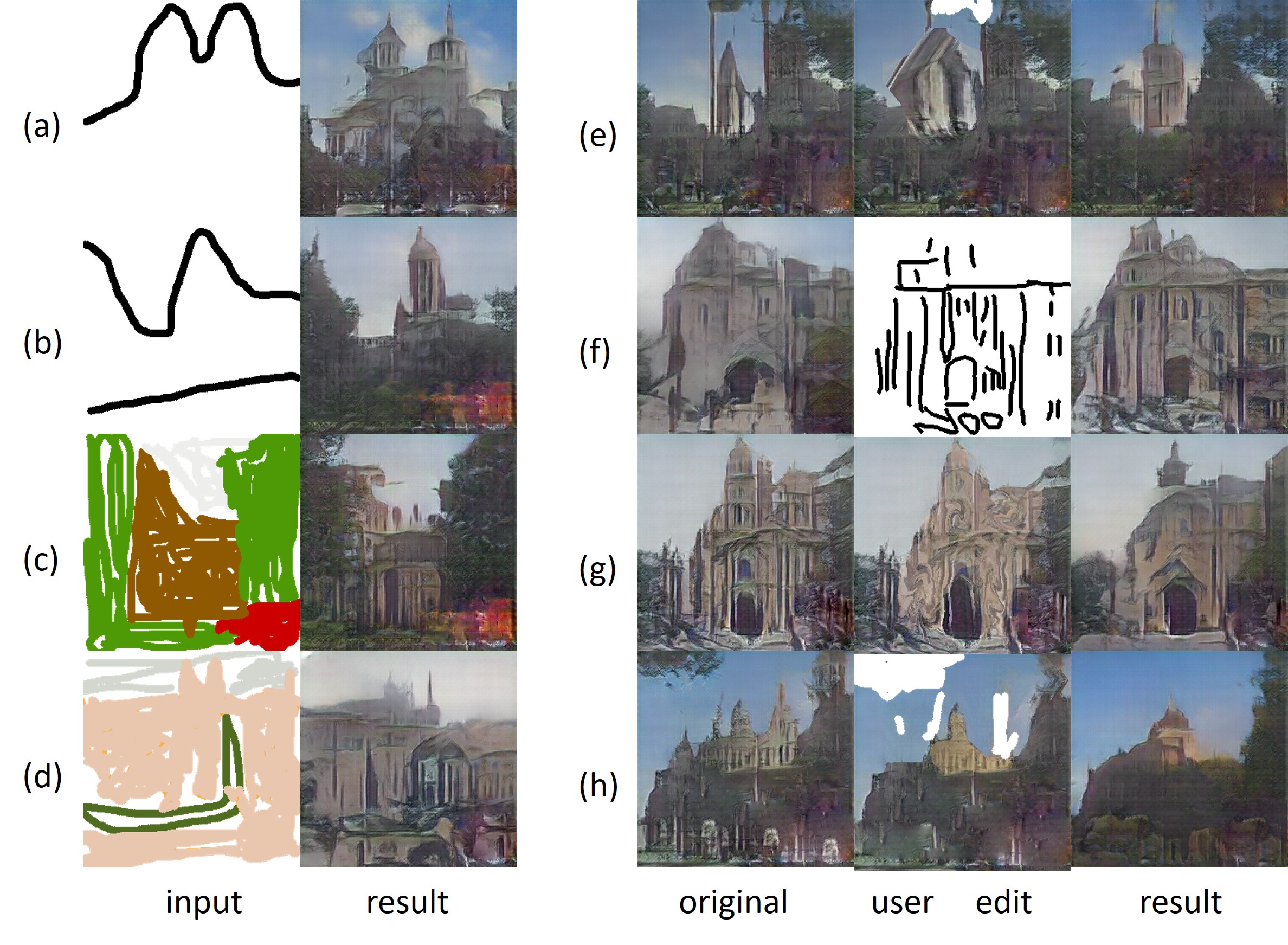}
\vspace{-10pt}

\caption{Church image generation and editing results with improved iGAN: (a-b) results based on edge maps; (c-d) results based on masked color maps; and (e-h) image editing results.} \label{fig:lsun-igan}
\end{center}
\vspace{-10pt}

\end{figure}

Figures \ref{fig:face256}, \ref{fig:car-igan} and \ref{fig:lsun-igan} show more image editing results with the modified iGAN, by manipulating the BSD-GAN manifolds trained on CelebA\_HQ \cite{karras2017progressive}, car and LSUN \cite{yu15lsun} church\_outdoor datasets separately.

\noindent

%% file: conclusion.tex
\section{Conclusion, limitation, and future work}
\label{sec:future}


We have introduced BSD-GAN, a novel, progressive training procedure for unconditional GANs which enables multi-scale image representation learning and manipulation. The key idea is to not only progressively increase network depth by adding layers, but also increase the network width by creating multiple, progressively activated training branches triggered by different sub-vectors of the network input. Each sub-vector corresponds to, and is trained to learn, image representations at a particular scale, leading to a scale-disentangled learning scheme.
Experimental results on several well-known high-quality image datasets verify the effectiveness of our method in disentangling image representations by scales. We also demonstrated new and improved applications via BSD-GAN.

BSD-GAN is scale-aware, but not feature-aware. This is a major limitation to progressive training using images at a selected set of resolutions since not all image features are well represented in this selected set of training images. Also, similar or repeated features in an image may not always be in the same scale, e.g., due to perspective projection. While such features are often manipulated as a collection during editing, they are difficult to learn using our current network. In addition, the scale disentanglement provided by BSD-GAN is only partial, since adding a new training layer corresponding to a newly activated sub-vector can still impact weights learned for the preceding sub-vectors. As a result, all learned weights may be correlated with image features across multiple scales. Overall, while BSD-GAN certainly improves controllability for image manipulation, it still has inherent limitations. 

In future work, we would like to extend BSD-GAN to feature- or semantic-aware progressive training, where sub-vector designation can be based on more meaningful or more visually apparent image features; this would add more meaning to sub-vector manipulation for image editing and synthesis. We believe that the progressive training paradigm introduced by BSD-GAN can be tuned for different forms of disentanglement by adjusting the training targets. In addition, we shall explore potential values of scale-disentangled image representations in tasks such as image compression, filtering, and denoising. Finally, an intriguing question is whether branch suppression exists in other ``multi-branch'' networks, e.g., ResNet~\cite{he2016deep}, DenseNet~\cite{huang2017densely}, and capsule networks~\cite{hinton2011transforming}. We are interested in whether this phenomenon may offer insights to the training of other convolutional and generative networks.